\documentclass[sigconf]{acmart}
\AtBeginDocument{%
  }

\copyrightyear{2025}
\acmYear{2025}
\setcopyright{acmlicensed}\acmConference[KDD '25]{Proceedings of the 31st ACM SIGKDD Conference on Knowledge Discovery and Data Mining V.1}{August 3--7, 2025}{Toronto, ON, Canada}
\acmBooktitle{Proceedings of the 31st ACM SIGKDD Conference on Knowledge Discovery and Data Mining V.1 (KDD '25), August 3--7, 2025, Toronto, ON, Canada}
\acmDOI{10.1145/3690624.3709203}
\acmISBN{979-8-4007-1245-6/25/08}

\usepackage{makecell}
\usepackage{graphicx}
\usepackage{subfigure}
\usepackage{colortbl}
\usepackage{algorithmic}
\usepackage[ruled,linesnumbered,vlined]{algorithm2e}
\usepackage{enumitem}
\usepackage{verbatim}
\usepackage{multirow}
\usepackage{stfloats}
\newcommand{\Mat}[1]{\mathbf{#1}}

\newcommand{\Set}[1]{\mathcal{#1}}

\newcommand{\ie}{\textit{i.e., }}
\newcommand{\eg}{\textit{e.g., }}

\definecolor{mygray}{gray}{0.6}
\definecolor{syd_color}{rgb}{0.858, 0.188, 0.478}

\usepackage{MnSymbol}

\def\eqref#1{equation~(\ref{#1})}
\def\Eqref#1{Equation~(\ref{#1})}

\newtheorem{proposition}{Proposition}
\newtheorem{lemma}{Lemma}

\theoremstyle{plain}

\newtheorem{definition}{Definition}

\newtheorem{assumption}{Assumption}

\theoremstyle{remark}

\begin{document}

%%
%% The "title" command has an optional parameter,
%% allowing the author to define a "short title" to be used in page headers.
\title{A Unified Invariant Learning Framework for Graph Classification}

%%
%% The "author" command and its associated commands are used to define
%% the authors and their affiliations.
%% Of note is the shared affiliation of the first two authors, and the
%% "authornote" and "authornotemark" commands
%% used to denote shared contribution to the research.
\author{Yongduo Sui}
% \authornote{This work was done during author’s internship at Ant Group.}
\email{yongduosui@tencent.com}
\affiliation{%
  \institution{Tencent}
  \city{Shenzhen}
  \country{China}}
  
% \author{Yongduo Sui}
% % \authornote{This work was done during author’s internship at Ant Group.}
% \email{syd2019@mail.ustc.edu.cn}
% \affiliation{%
%   \institution{University of Science and Technology of China}
%   \country{}}

\author{Jie Sun}
\email{sunjie2019@mail.ustc.edu.cn}
\affiliation{%
  \institution{University of Science and Technology of China}
  \city{Hefei}
  \country{China}}

\author{Shuyao Wang}
\email{shuyaowang@mail.ustc.edu.cn}
\affiliation{%
  \institution{University of Science and Technology of China}
  \city{Hefei}
  \country{China}}

\author{Zemin Liu}
\email{liu.zemin@zju.edu.cn}
\affiliation{%
  \institution{Zhejiang University}
  \city{Hangzhou}
  \country{China}
}

\author{Qing Cui}
\email{cuiqing.cq@antgroup.com}
\affiliation{%
  \institution{Ant Group}
  \city{Beijing}
  \country{China}
}

\author{Longfei Li}
\email{longyao.llf@antgroup.com}
\affiliation{%
  \institution{Ant Group}
  \city{Hangzhou}
  \country{China}
}

\author{Xiang Wang}
\authornote{Corresponding author. Xiang Wang is also affiliated with Institute of Artificial Intelligence, Institute of Dataspace, Hefei Comprehensive National Science Center.}
\email{xiangwang1223@gmail.com}
\affiliation{\institution{University of Science and Technology of China}
  \city{Hefei}
  \country{China}
}

\renewcommand{\shortauthors}{Yongduo Sui et al.}
%%
%% By default, the full list of authors will be used in the page
%% headers. Often, this list is too long, and will overlap
%% other information printed in the page headers. This command allows
%% the author to define a more concise list
%% of authors' names for this purpose.
% \renewcommand{\shortauthors}{Trovato et al.}

%%
%% The abstract is a short summary of the work to be presented in the
%% article.
\begin{abstract}
Invariant learning demonstrates substantial potential for enhancing the generalization of graph neural networks (GNNs) with out-of-distribution (OOD) data. 
It aims to recognize stable features in graph data for classification, based on the premise that these features causally determine the target label, and their influence is invariant to changes in distribution. 
Along this line, most studies have attempted to pinpoint these stable features by emphasizing explicit substructures in the graph, such as masked or attentive subgraphs, and primarily enforcing the invariance principle in the semantic space, \ie graph representations.
However, we argue that focusing only on the semantic space may not accurately identify these stable features.
To address this, we introduce the Unified Invariant Learning (UIL) framework for graph classification. 
It provides a unified perspective on invariant graph learning, emphasizing both structural and semantic invariance principles to identify more robust stable features.
In the graph space, UIL adheres to the structural invariance principle by reducing the distance between graphons over a set of stable features across different environments.
Simultaneously, to confirm semantic invariance, UIL underscores that the acquired graph representations should demonstrate exemplary performance across diverse environments.
We present both theoretical and empirical evidence to confirm our method's ability to recognize superior stable features. Moreover, through a series of comprehensive experiments complemented by in-depth analyses, we demonstrate that UIL considerably enhances OOD generalization, surpassing the performance of leading baseline methods. 
Our codes are available at https://github.com/yongduosui/UIL.
\end{abstract}


%%
%% Keywords. The author(s) should pick words that accurately describe
%% the work being presented. Separate the keywords with commas.
% \keywords{Do, Not, Us, This, Code, Put, the, Correct, Terms, for,
%   Your, Paper}
%% A "teaser" image appears between the author and affiliation
%% information and the body of the document, and typically spans the
%% page.
% \begin{teaserfigure}
%   \includegraphics[width=\textwidth]{sampleteaser}
%   \caption{Seattle Mariners at Spring Training, 2010.}
%   \Description{Enjoying the baseball game from the third-base
%   seats. Ichiro Suzuki preparing to bat.}
%   \label{fig:teaser}
% \end{teaserfigure}

% \received{20 February 2007}
% \received[revised]{12 March 2009}
% \received[accepted]{5 June 2009}

%%
%% This command processes the author and affiliation and title
%% information and builds the first part of the formatted document.
\maketitle

\begin{figure}[t]
\centering
\includegraphics[width=0.9\linewidth]{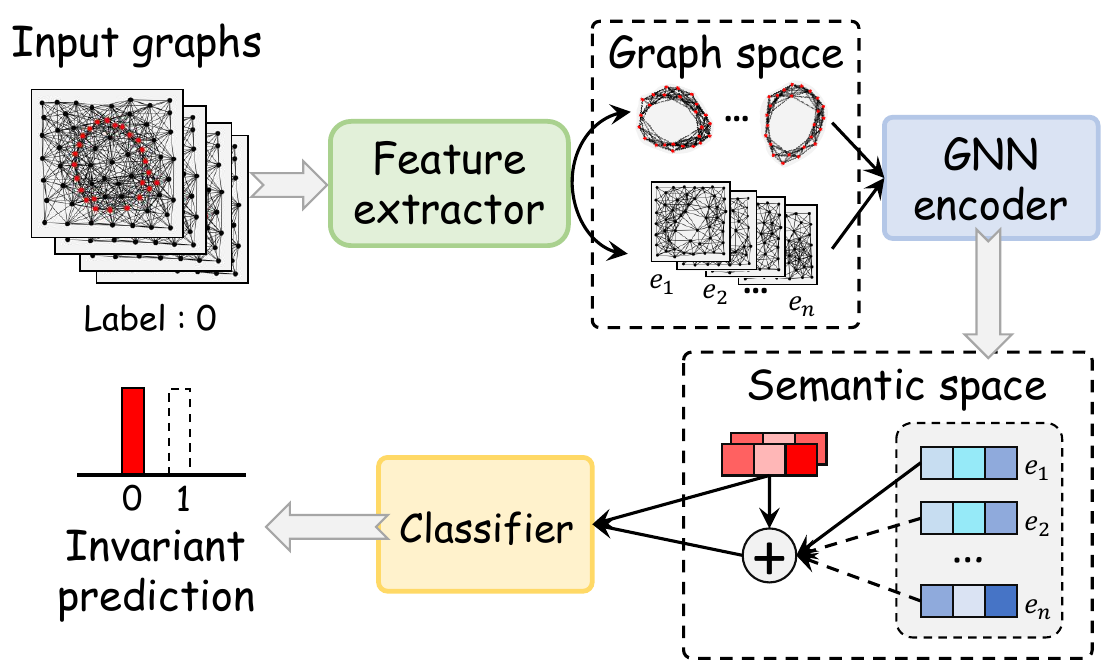}
\vspace{-4mm}
\caption{The workflow of several prevalent methods grounded in invariant graph learning.}
% including DIR \cite{DIR}, GREA \cite{liu2022graph}, CAL \cite{sui2021causal}, and DisC \cite{fan2022debiasing}. 
% These methods primarily strive to enforce the invariance principle within the semantic (\ie graph representation) space, aiming for feature extractors to accurately isolate stable features in the graph space.
% For the input graph data, red subgraph represents the stable features and black subgraph represents the environmental features.
\label{fig:teaser}
\vspace{-8mm}
\end{figure}

\section{Introduction}

Graph neural networks (GNNs) \cite{kipf2016semi,xu2018how,velivckovic2018graph} have achieved promising results on various graph classification tasks, such as social network analysis and molecular property prediction.
Such superior performance heavily relies on the assumption that training and test data are independently drawn from an identical distribution.
Unfortunately, recent studies \cite{li2022out,gui2022good} have shown that GNNs usually suffer from severe performance drops when facing the out-of-distribution (OOD) issue \cite{shen2021towards,ye2022ood}, where the training and test graphs are drawn from different distributions.
The poor generalization of GNNs hinders their practical deployment on critical applications.

Aiming at combating the distributional divergence between the training and test data, invariant learning \cite{arjovsky2019invariant,chang2020invariant} is provoking great interest in graph classification tasks \cite{DIR,sui2021causal,liu2022graph,li2022learning,fan2022debiasing}.
Specifically, for a graph being classified, it aims to discern a subset of features stable to the distributional uncertainty in training data.
Such stable features are usually based on two assumptions:
(1) they causally determine the target classification label \cite{DIR,fan2022debiasing}; and (2) their relationship with the label is invariant, regardless of distribution changes \cite{sui2021causal,li2022learning}.
In stark contrast, the complementary parts of stable features, termed the changing or environmental features \cite{liu2022graph}, are assumed to have no cause-effect relations with the label and often indicate the changes in data distribution.
In line with invariant learning, prevalent efforts \cite{DIR,liu2022graph,sui2021causal,fan2022debiasing} aim to identify and utilize stable features within the graph data to enhance generalization. 
As depicted in Figure \ref{fig:teaser}, a feature extractor network is commonly employed to distinctly segregate stable and environmental subgraphs within the graph space, allowing GNN models to leverage the extracted stable subgraphs for robust predictions.
% As illustrated in Figure \ref{fig:teaser}, they typically employ a feature extractor network to distinctly segregate stable and environmental subgraphs within the graph space. The intent is for the GNN models to leverage the extracted stable subgraphs to generate robust predictions.
To pinpoint these stable subgraphs, most approaches adhere to the invariance principle, which stipulates that given stable features, the classifier's predictions should remain invariant despite changes in environmental features. However, synthesizing ``counterfactual graphs'' directly within the graph space presents challenges due to the non-Euclidean nature of graph data \cite{sui2021causal}. 
Consequently, the invariance principle is often enforced within the semantic (\ie graph representation) space \cite{volpi2018generalizing}, as illustrated in Figure \ref{fig:teaser}. This strategy encourages GNN models to learn environment-invariant representations, thereby enhancing the feature extraction network's capacity to identify superior stable features within the graph space.
% As a result, they enforce the invariance principle within the semantic (\ie representation) space, as shown in Figure \ref{fig:teaser}. This encourages the GNN model to learn an environment-invariant representation and further promotes the feature extraction network's ability to capture better stable features in the graph space.

Regrettably, our findings indicate that most existing invariant learning paradigms do not necessarily ensure the feature extractor's ability to effectively identify what we term ``minimal stable features''. These are defined as oracle rationale \cite{DIR} that uniquely and causally determines the label, maintaining an invariant relationship with the label across all training and test distributions. 
To illustrate, in the context of molecular graph property prediction, the chemical properties of molecules are causally determined by their functional groups. For instance, the ``-COOH'' group determines acidity, while ``-NH$_2$'' influences reducibility. These can be considered examples of minimal stable features \cite{hu2020open}.
However, given the limited scale of training data collection, a specific subset of environmental features, termed ``stable-environmental features'', may emerge in training data. These features bear no causal relationship with labels, and they respond insensitively to alterations across diverse training environments. Thus, confining the implementation of invariance solely to the semantic space runs the risk of misclassifying these trivial features as stable features. If these trivial features demonstrate instability outside the training data, the model's generalization capability cannot be assured.
% However, certain distinct environmental features in the training data, referred to as ``stable-environmental features'', may not sensitively mirror changes in diverse environments due to the limited scale of the training data collection. 
% \zeminC{上文的stable environment features不太明确是指什么，可以举例简单来说明一下。或者利用II.A中标蓝色的那句话来解释下也行。}
% Given that these methodologies primarily emphasize the invariance of the estimated stable features across diverse training environments within the semantic space, they run the risk of misclassifying stable-environmental features as stable features. Yet, these trivial features may lack stability outside the training distribution, thereby compromising the model's generalization capabilities. 
% \zeminC{这里最好简单解释一下，为什么在training之外比较少，可以举一个或者多个这种场景的例子。}
Therefore, it is essential to re-design the prevailing invariant learning paradigm to rectify these issues.

To tackle this challenge, we reassess the distinction between minimal stable features and other environmental features. Typically, there are notable structural resemblances among the minimal stable features within a class of population. For instance, different molecules with identical labels often contain functional groups that follow a recurring or consistent pattern. To illustrate, acidic molecules tend to present enduring ``-COOH'' based patterns. In the context of MNIST superpixel graphs \cite{knyazev2019understanding}, a class of graphs typically exhibits a similar pattern, evident in the presence of comparable handwritten digit patterns within each class of graphs. As such, these patterns, serving as the embodiment of class-specific knowledge, implicitly pervade each data class. In contrast, achieving class-related structural consistency or invariance within the graph space proves to be difficult due to the absence of a causal link between other environmental features and labels \cite{DIR}. Such observations prompt us to pose a critical question: \emph{Is it possible to impose additional constraints on the graph space, thus ensuring that the model accurately identifies minimal stable features?}

In this paper, we advocate that graph learning should simultaneously consider both structural and semantic invariance. To this end, we propose a Unified Invariant Learning (UIL) Framework aimed at extracting minimal stable features for improved generalization.
Specifically, in the graph space, we draw inspiration from graphon --- a function effective at capturing generative patterns within a series of graphs. We convert the class-related structural invariance, implicitly present in each class of graph data, into explicit graphons. 
By promoting the invariance of these graphons across different environments, our model can effectively identify minimal stable features, thereby ensuring structural invariance.
In the semantic space \cite{volpi2018generalizing}, we motivate models to learn invariant graph representations across varying environments.
% In the semantic space, we motivate models to uphold invariant predictions across varying environments, predicated on the estimated stable features they capture.
% \zeminC{这里需要解释一下为什么称为semantic space，好像未体现出semantic？可以一句话简单解释一下，上文中解释比较少。}
Consequently, UIL proffers a unified perspective on invariant learning, incorporating both structural and semantic invariance. 
It fosters the identification of minimal stable features, thereby bolstering the model's generalization ability.
The main contributions are summarized as follows:

\begin{itemize}[leftmargin=4mm]
% \item We expose the limitations of current invariant graph learning paradigms, which emphasize invariance solely within the semantic space and often struggle to capture minimal stable features in the graph space.
% \item We uncover the limitations of most invariant graph learning paradigms, which focus solely on semantic invariance and often fail to capture minimal stable features accurately. \zeminC{这一点contribution，最好不是直接喷别人的flaws，最好是这样的句式：我们发现别人的XX问题，然后提出从YY角度（只说high-level的角度/idea，不用说具体的model）来解决（如，we reassess the distinction between minimal stable features and other environmental features）。}
\item We underscore the pivotal role of imposing structural invariance within the graph space in the context of invariant graph learning. This strategy significantly aids the model in capturing minimal stable features.
\item We present UIL, a graph learning framework that emphasizes both structural and semantic invariance, thereby enhancing the model generalization.
\item We provide both theoretical and empirical evidence to support our structural invariance's role in learning minimal stable features. Detailed comparative results further highlight the superiority of our approach.
% \item We offer theoretical and experimental proofs supporting our structural invariance's role in learning minimal stable features. Extensive comparison results further underscore the superiority of our approach.
\end{itemize}

\section{Preliminaries}

% In this section, we first give the notations used in this paper, then introduce the definitions of causal and environmental features, and finally we rethink the essence of causal features from a global view in data distribution.

In this paper, we adopt a lowercase letter (\eg $g$) to describe an instance, the same uppercase letter (\eg $G$) to denote a random variable, the calligraphic font (\eg $\Set{G}$) to define a set, and the blackboard bold typeface (\eg $\mathbb{G}$) to define a space.
We denote a graph by $g=\{\Mat{A}, \Mat{X}\}$ with node set $\Set{V}$ and edge set $\Set{U}$. 
$y \in \mathbb{Y}$ is a graph label and $C$ is the class number.
Let $\Mat{A}\in \mathbb{R}^{|\mathcal{V}|\times |\mathcal{V}|}$ denote the adjacency matrix, where $\Mat{A}[i,j]=1$ if edge $(v_i, v_j) \in \Set{U}$, otherwise $\Mat{A}[i,j]=0$. 
Let $\Mat{X}\in \mathbb{R}^{|\mathcal{V}|\times d_x}$ be the node feature matrix, where $\Mat{x}_i=\Mat{X}[i,:]$ is a $d$-dimensional feature vector of node $v_i \in \Set{V}$. 
We define $W$ as a graphon.
The important notations are summarized in Table \ref{table:notations}.

\begin{table}[t]
\centering
\caption{\centering Summary of notations and descriptions.}
% \vspace{-2mm}
\label{table:notations}
\begin{tabular}{c|c}
\hline
Notations & Descriptions \\
\hline
% \hline
$g, G, \Set{G}, \mathbb{G}$	     & The graph instance, random variable, set, and space. \\
$y, Y, \Set{Y},\mathbb{Y} $	     & The label instance, random variable, set, and space. \\
$v_i, \Set{V}$	                 & The node $v_i$ and node set $\Set{V}$, $v_i\in\Set{V}$. \\
$(v_i, v_j), \Set{U}$	         & The edge $(v_i, v_j)$ and edge set $\Set{U}$, $(v_i, v_j)\in\Set{U}$. \\
$\Mat{A}$	                     & The adjacency matrix $\Mat{A}\in \{0,1\}^{|\Set{V}|\times |\Set{V}|}$. \\
$\Mat{X}$	                     & The node feature matrix $\Mat{X}\in \mathbb{R}^{|\Set{V}|\times d_x}$. \\
$\Mat{x}_i$	                     & The feature of node $v_i$, and $\Mat{x}_i \in \mathbb{R}^{d_x}$. \\
$\ell, \Set{L}$	                 & The loss function and the optimization objective. \\
$C $	                         & The class number. \\
$\Set{E}_{tr}, \Set{E}_{te}$	 & The training and test environments. \\
$\Set{D}_{tr}, \Set{D}_{te}$	 & The training and test sets. \\
$P_{tr}, P_{te}$	             & The training and test distributions. \\
$W, \mathbb{W}$	                 & The graphon \& step function and graphon space.\\
$\Mat{W}$	                     & The step function in matrix form.\\
$\Omega$	                     & The measurable space.\\
$\phi$	                         & The measure-preserving map, $\phi: \Omega\rightarrow\Omega$.\\
$||\cdot||_{\Box}$	             & The cut norm, and $||\cdot||_{\Box}:\mathbb{W} \rightarrow \mathbb{R}$. \\
$\delta_{\Box}(\cdot, \cdot)$	 & The cut distance, and $\delta(\cdot,\cdot):\mathbb{W}^2 \rightarrow \mathbb{R}$.\\
$\Mat{M}^x, \Mat{M}^a$	         & The masks for node features and adjacency matrix. \\
$\Mat{z}_i$	                     & The representation of node $v_i$, and $\Mat{z}_i \in \mathbb{R}^{d_z}$. \\
$\Mat{Z}$	                     & The node representation matrix $\Mat{Z} \in \mathbb{R}^{|\Set{V}|\times d_z}$. \\
$\Mat{h}, \Set{H}$	             & The graph representation $\Mat{h} \in \mathbb{R}^{d_h}$ and set. \\
$f_{\theta}, \Theta$	         & The model $f:\mathbb{G}\rightarrow\mathbb{Y}$ parameterized by $\theta\in\Theta$. \\
$\Phi$, $\Psi$                   & The feature extractor $\Phi: \mathbb{G}\rightarrow\mathbb{G}$, and MLP network. \\
$h, w$	                         & The graph encoder and classifier. \\
$\odot$	                         & The element-wise (Hadamard) product. \\
\hline
\end{tabular}
\vspace{-2mm}
\end{table}

\subsection{OOD Issue in Graph Classification}
In this work, we focus on the graph classification task.
Specifically, we define a training set $\Set{D}_{tr}=\{(g_i, y_i)\}$ drawn from training distribution $P_{tr}(G,Y)$.
Given a loss function $\ell$ and a parameter space $\Theta$, we need to train a model $f_\theta: \mathbb{G} \rightarrow \mathbb{Y}$ on $\Set{D}_{tr}$. 
Existing GNNs \cite{kipf2016semi,xu2018how} adopt the Empirical Risk Minimization (ERM) to optimize the following objective:
\begin{equation}\label{equ:erm} 
    \theta^*:=\mathop{\arg\min}_{\theta\in\Theta}\mathbb{E}_{(g,y)\sim D_{tr}}[\ell(f_\theta(g), y)].
\end{equation}
Then we need to use model $f_{\theta^*}$ to infer labels in test set $\Set{D}_{te}$, where the data is sampled from the test distribution $P_{te}(G,Y)$.
If $P_{tr}(G,Y)\neq P_{te}(G,Y)$, the OOD issue will occur, and the performance of the GNN model will drop sharply.
% The OOD problem in graph classification is very common in practice.
The OOD problem is widespread in graph classification tasks.
% For molecular property prediction, we usually need to train models with past datasets and hope that the models can predict molecules from different distributions in the future.

To investigate the cause of the OOD issue, we revisit the graph generation mechanism.
Following the commonly-used assumption of graph generation \cite{DIR,sui2021causal,liu2022graph}, there exist stable features ${G}_{st}$ in graph data, which causally determine the labels.
These features are often called rationales \cite{DIR}, causal subgraphs \cite{fan2022debiasing} or causal features \cite{sui2021causal,sui2023unleashing} in the existing literature.
The relationship between stable features and labels is assumed to be invariant across distributions.
While their complementary parts ${G}_{en}$, often called environmental features \cite{liu2022graph,sui2023unleashing}, have no causal-effect with labels and often indicate the changes in data distribution.
For given training data and test data, they may contain data subsets from multiple environments, \ie $\Set{D}_{tr}=\{\Set{G}^{e_i} | e_i \in \Set{E}_{tr}\}$ and $\Set{D}_{te}=\{\Set{G}^{e_i} | e_i \in \Set{E}_{te}\}$, where $\Set{G}^{e_i}$ denote a group of graphs under the environment $e_i$. 
And we define $\Set{E}$ as the set of all possible environments, so we have $\Set{E}=\Set{E}_{tr} \cup \Set{E}_{te}$.
Hence, the instability of the environmental features often leads to various distribution shifts \cite{gui2022good}.

% Formally, we define these two features as follows:
% \begin{assumption}[Graph Data Generation]\label{ass:1}
% Given a graph $G$, the stable feature ${G}_{st}$ and environmental feature ${G}_{en}$ satisfy the following conditions:
% (1) Sufficiency condition: The stable feature preserves the core information of $G$ related to the label $Y$. This is formally expressed as $P({Y|G_{st}}) = P(Y|G)$.
% (2) Independence condition: The label $Y$ is independent of the environmental feature, conditioned on the stable feature. This is expressed as $Y \upmodels G_{en} \mid G_{st}$.
% \end{assumption}
% It's important to emphasize that Assumption 1 corresponds to the data generation process, a process that always holds true across all distributions, encompassing both training and test distributions. 
% Consequently, a straightforward method to achieve OOD generalization involves striving to learn stable features, $G_{st}$, whose relationships with labels remain invariant across both the training and test distributions \cite{ye2022ood}.

% Consequently, a straightforward method to attain OOD generalization involves endeavoring to learn stable features $G_{st}$ that are common to both the training and test distributions.

\subsection{Invariant Learning for Graph Classification}
% Invariant graph learning has emerged as a popular strategy to capture stable features in graph data and improve the generalization capabilities of GNN models.  
Invariant graph learning, originally derived from Invariant Risk Minimization (IRM) \cite{arjovsky2019invariant} within the realm of Euclidean data, attains OOD generalization by learning graph representations invariant to environmental changes. However, due to the non-Euclidean nature of graph data, directly transferring the penalty term from IRM to graph learning poses a challenge, often leading to suboptimal performance \cite{DIR,gui2022good}. Consequently, a burgeoning line of research focuses on extracting stable subgraphs from the graph space to achieve invariant learning \cite{DIR,liu2022graph,sui2021causal,li2022learning,fan2022debiasing}.
The fundamental framework is illustrated in Figure \ref{fig:teaser}. This architecture mainly comprises three components: the feature extractor network $\Phi$, GNN encoder $h$, and classifier $w$. The core idea is to extract stable features via the feature extractor, subsequently inputting these features into the GNN encoder and classifier for predictions. If the feature extractor can effectively capture stable features $G_{st}$ in graph data across all possible environments, then the model can achieve excellent generalization under distribution shift. To ensure invariance across environments, these methods generally uphold the semantic invariance principle.

\begin{definition}[Semantic Invariance Principle]\label{def:1}
Given a graph $G$ and its label $Y$, an optimal model $f^*$ should satisfy:
(1) Sufficiency condition: $Y=\omega^*(h^*(\Phi^*(G)))+\epsilon, \epsilon \upmodels G$, where $\epsilon$ is random noise.
(2) Invariance condition: $P^e(Y|\Phi^*(G))=P^{e'}(Y|\Phi^*(G)), \forall e, e' \in \Set{E}$.
\end{definition}

Given the features extracted by $\Phi$, if $Y$ can be completely determined and $P^e(Y|\Phi^*(G))=P^{e'}(Y|\Phi^*(G))$ in every environment, the invariance principle is thus fulfilled. 
% It reveals that the extracted features do indeed meet the principle to be classified as a stable subgraph. 
However, it's crucial to note that the environment stipulated in Definition \ref{def:1} corresponds to the training environment (\ie $\Set{E}=\Set{E}_{tr}$) in practice. 
It implies that the stable features identified by the semantic invariance might surpass the scope of the grouth-truth stable features. In light of this, we provide the subsequent definitions.

\begin{definition}[Estimated Stable Feature]\label{def:est}
Given the training distribution $P_{tr}(G,Y)$, let $\hat{G}_{st}$ denote any subset (\eg subgraph) of the graph data $G$, \ie  $\hat{G}_{st} \subseteq G$. If $\hat{G}_{st}$ and $G$ satisfy the condition: $\mathbb{E}_{P_{tr}}[Y|\hat{G}_{st}]=\mathbb{E}_{P_{tr}}[Y|G]$, then $\hat{G}_{st}$ is referred to as an estimated stable feature within the training data. 
\end{definition}
% \zeminC{这里的graph data这个词有点抽象，能否举个例子，如such as a subgraph.}

We hereby define the set of all estimated stable features, as specified in Definition \ref{def:est} for $Y$, as ${\rm Stable}_{P_{tr}}(Y)$. For the sake of simplicity, we will represent the set under the training distribution $P_{tr}$ as ${\rm Stable}(Y)$, that is ${\rm Stable}(Y)\triangleq{\rm Stable}_{P_{tr}}(Y)$.

% Now we can define the set of all stable features in Definition \ref{def:est} for $Y$ as ${\rm Stable}_{P_{tr}}(Y)$. In addition, we use ${\rm Stable}(Y)$ to denote the set under the training distribution $P_{tr}$ for simplicity, \ie  ${\rm Stable}(Y)={\rm Stable}_{P_{tr}}(Y)$.

\begin{definition}[Minimal Stable Feature Set]\label{def:3}
Given the training distribution $P_{tr}(G,Y)$, we can choose a subset ${\rm Stable}'(Y)\subseteq {\rm Stable}(Y)$, if $\forall G' \subset{\rm Stable}'(Y)$ and $\mathbb{E}_{P_{tr}}[Y|G']\neq\mathbb{E}_{P_{tr}}[Y|G]$, then we define ${\rm Stable}'(Y)={\rm MinStable}(Y)$ as the minimal stable feature set in the training data.
\end{definition}

From Definitions \ref{def:est} and \ref{def:3}, we notice that ${\rm MinStable}(Y)$ equates to $G_{st}$ in the data generation process. 
% These features typically fall within the ${\rm Stable}(Y)$ in most instances. 
Most strategies, constrained by limitations in training environments, can only discern ${\rm Stable}(Y)$ within the training data based on the invariance principle. Under such circumstances, the model may assimilate certain ``stable-environmental features'', here denoted as $G_{se}:=\hat{G}_{st} \backslash G_{st}$. 
These features might display instability beyond the training distribution, thereby compromising the generalization capability.
Consequently, it may be beneficial to consider introducing additional invariance constraints within the graph space. Such constraints could assist in further differentiating these trivial features from ${\rm MinStable}(Y)$.

\section{Methodology}
This section provides our detailed implementations of the UIL.
The overview of the UIL framework is depicted in Figure \ref{fig:model}. 
Specifically, upon receiving the input graphs, the feature separator initially procures the estimated stable and environmental features (Section \ref{sec:sta_env}). On the one hand, we construct distance constraints for a group of estimated stable features within the graphon space to guarantee structural invariance (Section \ref{sec:graphon}). On the other hand, we acquire invariant graph representations within the representation space to impose semantic invariance (Section \ref{sec:sem}). 

\subsection{Stable \& Environmental Feature Separator}\label{sec:sta_env}
We define our model as $f=\omega \circ h \circ \Phi$, which includes three modules $\Phi$, $h$ and $\omega$.
$\Phi(\cdot)$ is a stable feature extractor that targets stable features; $h(\cdot)$ is a feature encoder network (with readout function), which can obtain graph-level representation; $\omega(\cdot)$ is a classifier.
Given a graph $g$ with node set $\Set{V}$ and edge set $\Set{U}$, $\Phi$ first obtains the node representation $\Mat{Z}\in \mathbb{R}^{|\Set{V}|\times d_z}$ via a GNN-based encoder ${\rm GNN}(\cdot)$, denoted as $\Mat{Z}={\rm GNN}(g)$.

Then we utilize two MLP networks $\Psi_1$ and $\Psi_2$ to generate stable masks, which denote the importance of nodes or edges to capture the stable features in graphs.
Specifically, we can generate stable masks for node $v_i \in \Set{V}$ and edge $(v_i, v_j) \in \Set{U}$:
\begin{equation}\label{equ:causaler} 
\Mat{M}_{i}^x=\sigma(\Psi_1(\Mat{z}_i)), \;\; \Mat{M}^a_{ij}=\sigma(\Psi_2([\Mat{z}_i, \Mat{z}_j])),
\end{equation}
where $\Mat{M}_{i}^x$ and $\Mat{M}^a_{ij}$ are the values of the node mask matrix $\Mat{M}^x\in \mathbb{R}^{|\Set{V}|\times 1}$ and edge mask matrix $\Mat{M}^a\in \mathbb{R}^{|\Set{V}\times |\Set{V}|}$, respectively; $\Mat{z}_i=\Mat{Z}[i,:]$ refers to the representation of node $v_i$; $\sigma$ is sigmoid function that projects the mask values to $[0,1]$. 
% Since the environmental feature is the complementary part of the stable feature, 
We define the environmental mask as the complement of the stable mask.
Hence, we can generate environmental masks to capture environmental features from nodes: $\overline{\Mat{M}}^x=\Mat{1}-\Mat{M}^x$ and edges: $\overline{\Mat{M}}^a=\Mat{1}-\Mat{M}^a$, respectively, where $\Mat{1}$ is the all-one matrix.
Finally, we can adopt these masks to estimate the stable feature $\hat{g}_{st}=\{\Mat{A}\odot\Mat{M}^a, \Mat{X}\odot\Mat{M}^x\}$ and environmental feature $\hat{g}_{en}=\{\Mat{A}\odot\overline{\Mat{M}}^a, \Mat{X}\odot\overline{\Mat{M}}^x\}$, where $\odot$ is element-wise multiplication.

\subsection{Stable Graphon Estimation}\label{sec:graphon}

We extract and summarize the implicit class-wise knowledge with stable generative patterns from a group of graph data with the same labels. We apply the graphon (in Definition \ref{def:graphon}) to the stable features, and give the definition of stable graphon as follows.

\begin{definition}[Stable Graphon]\label{def:sg}
Stable graphon is a graph function that can explicitly instantiate the class-specific pattern across a class of graphs.
Specifically, given a group of graphs $\Set{G}=\{g_i\}$ with the same label $y$, let $\Set{G}_{st}=\{g_{st}\}$ denote their corresponding stable features.
We can use these stable features to estimate a graphon: $\Set{G}_{st} \Rightarrow W_{st}$, and we define $W_{st}$ as the stable graphon of the graph set $\Set{G}$. 
\end{definition}
We can observe that the stable graphon explicitly defines the implicit class-specific knowledge in group of graph data.

\textbf{Graphon Estimation.}
In the actual implementation, it is difficult to directly obtain closed-form solutions of graphons.
Hence, we follow the commonly-used step function \cite{xu2021learning,han2022g}, which can be formulated as a matrix, to estimate the graphon.
Specifically, a step function $W_{\Set{P}}:\Omega^2\rightarrow[0,1]$ is an estimated graphon of a sequence of graphs, which can be expressed as 
$$W_{\Set{P}}(x,y)=\sum_{n,n'=1}^N w_{nn'}\mathbb{I}_{\Set{P}_n\times\Set{P}_{n'}}(x,y),$$
where $\Set{P}=(\Set{P}_1,..,\Set{P}_N)$ is an equitable partition of $\Omega$.
The value $w_{nn'}\in[0,1]$, and the indicator function $\mathbb{I}_{\Set{P}_n\times\Set{P}_{n'}}(x,y)$ is 1 if $(x,y)\in\Set{P}_n\times\Set{P}_{n'}$, otherwise it is 0. 
For simplicity, we rewrite the step function $W_{\Set{P}}$ as a $N\times N$ matrix $\Mat{W}_{\Set{P}}=[w_{nn'}]\in[0,1]^{N\times N},n,n'\in[1,N]$. 
For an observed graph $g$ with adjacency matrix $\Mat{A}$, it can be represented by a step function 
$$\begin{aligned}
W_{\Set{P}}^g &= \frac{1}{N^2}\sum_{i,j=1}^{|\mathcal{V}|}\Mat{A}[i,j]\mathbb{I}_{\Set{P}_i\times\Set{P}_j}(x,y) \\
&= \sum_{n,n'=1}^N w_{nn'}\mathbb{I}_{\Set{P}_n\times\Set{P}_{n'}}(x,y),
\end{aligned}$$
with $N$ equitable partitions of $\Omega$, and rewrite as a $N\times N$ matrix $\Mat{W}_{\Set{P}}^g$.
Suppose $W_\mathcal{P}(x,y)=\sum_{n,n'=1}^N w_{nn'}\mathbb{I}_{\mathcal{P}_n\times\mathcal{P}_{n'}}(x,y)$ is the estimated step function of unknown graphon $W$. The weak regularity lemma in \cite{lovasz2012large} has shown that every graphon can be approximated well in the cut norm by step functions.
\begin{lemma}[Weak Regularity]\label{lemma:1}
For every graphon $W\in\mathbb{W}$ and partition $\Set{P}$ of $\Omega$, $|\Set{P}|=N\geq 1$, there always exists a step function $W_{\Set{P}}$ such that
$$||W-W_{\Set{P}}||_{\Box}\leq\frac{2}{\sqrt{\log N}}||W||_2.$$
\end{lemma}
Depending on Lemma \ref{lemma:1}, we want to learn a step function $W_{\Set{P}}$, such that the cut distance between the ground truth and $W_{\Set{P}}$ is minimized, \ie the oracle step function $W_{\Set{O}}$ should satisfy
$$W_\Set{O}=\arg\min_{W_\Set{P}}\delta_{\Box}(W,W_\Set{P}).$$
However, the ground truth graphon $W$ is unknown, so we turn to observed graphs. We align graphs by rearranging the indexes of nodes based on the edge mask of the stable feature separator, \ie align nodes by $\frac{1}{2}\left(\sum_i\mathbf{M}^a[i,j]+\sum_j\mathbf{M}^a[i,j]\right)$.
Then we obtain a group of well-aligned adjacency matrices $\Mat{\hat{A}}_m$ and well-aligned step functions $\hat{W}_{\Set{P}}^{g_m}$.
Since the observed graphs $\mathcal{G}$ have the similar topological pattern with ground truth graphon $W$, our graphon estimation method should minimize $\sum_{m=1}^M\delta_{\Box}(\hat{W}_{\Set{P}_m}^{g_m}, W_\mathcal{P})$, \ie the optimal step function $W_\Set{P}^*$ should satisfy
$$W_\Set{P}^* = \arg\min_{W_\Set{P}}\sum_{m=1}^M\delta_{\Box}(\hat{W}_{\Set{P}_m}^{g_m}, W_\mathcal{P}).$$
Thus we can get the optimal step function $W_{\Set{P}}^*$ by solving the above optimization problem. We use the methods in GWB algorithm \cite{xu2021learning} to estimate the cut distance between any two step functions.

\begin{figure*}[t]
\centering
\includegraphics[width=1\linewidth]{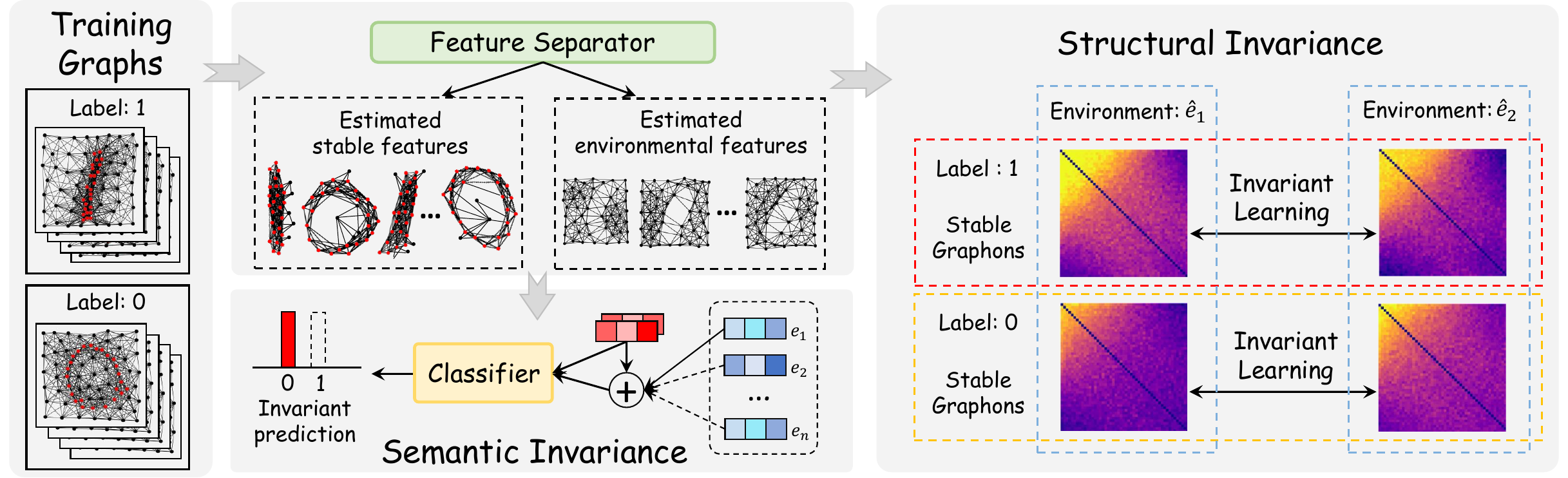}
\vspace{-6mm}
\caption{The overview of the proposed Unified Invariant Learning (UIL) framework.}
\label{fig:model}
\vspace{-4mm}
\end{figure*}

\subsection{A Unified View for Invariant Learning}\label{sec:sem}

We enforce the invariance constraints within the graph space, thus leading us to propose the principle of structural invariance.

\begin{definition}[Structural Invariance Principle]\label{def:structure}
Given a dataset $\Set{D}$ with environment set $\Set{E}$, we sample two graph sets $\Set{G}^e$ and $\Set{G}^{e'}$ with the same label $y$, where $\forall e, e' \in \Set{E}$.
Let $\Set{G}^e_{st}$ and $\Set{G}^{e'}_{st}$ denote the captured stable feature sets via feature extractor $\Phi$.
Then we can use these stable feature sets to estimate two stable graphons: $\Set{G}^e_{st} \Rightarrow W^e_{st}$ and $\Set{G}^{e'}_{st} \Rightarrow W^{e'}_{st}$.
An optimal stable feature extractor $\Phi^*$ should make the graphons satisfy: $\delta(W^{e}_{st},W^{e'}_{st})$ = 0, where $\delta(\cdot,\cdot):\mathbb{W}^2 \rightarrow \mathbb{R}$ is a distance function defined on the graphon space $\mathbb{W}$.
\end{definition}

Intuitively, our approach involves a more comprehensive introduction of inductive bias for invariant learning on graphs. Essentially, we assume that stable features of a class of graph data share the same generative pattern, which can be modeled by the same graphon. This generative pattern is assumed to be invariant across all possible distributions or environments, including unknown test distributions.
However, the strategy of imposing only semantic invariance does not introduce an inductive bias from the perspective of the graph space. Therefore, it can only yield features that maintain semantic invariance across training environments. These features, although stable within the training data, are not necessarily minimal stable features, making it challenging to ensure generalization to unknown test distributions.
Consequently, we aim to constrain directly within the graph data space, driving our model to capture features that are closer to the minimal stable feature. This holistic approach, encompassing both structural and semantic invariance, fosters a more robust method for the extraction of minimal stable features, facilitating improved generalization across diverse and unknown test distributions.

\textbf{Cut Distance Minimization.}
We utilize cut distance (in Definition \ref{def:cut}) to replace $\delta(\cdot,\cdot)$ in structural invariance principle, which can effectively measure the distance between two graphons.
Using the cut distance to constrain the distance of stable graphons across different environments, we can define the objective of structural invariant learning as:
\begin{equation}\label{equ:graphon_loss}
\hat{\mathcal{L}}_{str} = \sum_{y=1}^C\sum_{\hat{e}_i, \hat{e}_j\in\hat{\Set{E}}}\delta_{\Box}(W_{\hat{e}_i}^{y},W_{\hat{e}_j}^{y}),
\end{equation}
where $W_{\hat{e}_i}^{y}$ is the stable graphon of a group of graphs with label $y$ and the environment $\hat{e}_i$.
Furthermore, we derive Proposition \ref{the:cut} based on the properties of cut distance.
\begin{proposition}\label{the:cut}
Given graphons $W_1$ and $W_2$, their cut distance is bounded by $L_2$ norm:
$\delta_{\Box}(W_1,W_2)\leq ||W_1-W_2||_2$.
\end{proposition}
Since it is difficult to directly compute or minimize the cut distance \cite{gra_cut}, we adopt Proposition \ref{the:cut} to simplify the process of solving the cut distance.
Furthermore, we adopt the step function in matrix form $\Mat{W}$ to represent the graphon. 
Hence, we can rewrite \Eqref{equ:graphon_loss} as:
\begin{equation}\label{equ:graphon_loss2}
\mathcal{L}_{str} = \sum_{y=1}^C\sum_{\hat{e}_i, \hat{e}_j\in\hat{\Set{E}}}||\Mat{W}_{\hat{e}_i}^{y}-\Mat{W}_{\hat{e}_j}^{y}||_F,
\end{equation}
where $||\cdot||_F$ is the Frobenius norm, and we define \eqref{equ:graphon_loss2} as the objective of structural invariant learning.  It encourages the invariance of stable graphons across environments.

Semantic invariance chiefly serves to eliminate the unstable environmental features, which in turn fortifies the efficiency of structural invariance. Accordingly, our framework also seeks to ensure invariance within the representation space, adhering to semantic invariance principle. We begin by formulating the optimization objective for sufficiency:
\begin{equation}\label{equ:sufficiency} 
 \mathcal{L}_{sta} = \mathbb{E}_{(g_i, y_i)\sim \mathcal{D}_{tr}}[\ell(\omega(h(\hat{g}_{{st}_i})), y_i)],
\end{equation}
where $\ell$ denotes the cross-entropy loss.
Since stable features are part of the full graph, we also need to constrain the masks $\Mat{M}^a$ and $\Mat{M}^x$, to avoid trivial solutions.
Specifically, we define the following regularization objective:
\begin{equation}\label{equ:reg} 
    \Set{L}_{reg} = \mathbb{E}_{g\sim \mathcal{D}_{tr}}[r(\Mat{M}^x, |\Set{V}|,\rho) + r(\Mat{M}^a, |\Set{U}|,\rho)],
\end{equation}
where $r(\Mat{M},k,\rho)=(\sum_{ij}\Mat{M}_{ij}/k-\rho) + (\sum_{ij}\mathbb{I}[\Mat{M}_{ij}>0]/k-\rho)$; $\rho$ is stable ratio; $k$ is the total number of elements to be constrained; $\mathbb{I}\in\{0,1\}$ is an indicator function.
The first term encourages the average ratio close to $\rho$, while the second term encourages an uneven distribution.
For brevity, we define the sufficiency objective as: $\Set{L}_{suf}=\mathcal{L}_{sta}+\Set{L}_{reg}$.
It's important to note that we designate the ratio of stable features $\rho$ as a learnable parameter. This choice stems from two considerations:
1) We lack specific knowledge about the size of the minimal stable features, which makes it challenging to ascertain the proportion of stable features.
2) The potential existence of stable-environmental feature set, \ie $\hat{G}_{st} \backslash G_{st}$, necessitates the adaptive adjustment of the feature extractor through structural invariant learning. This ensures that the feature extractor ultimately converges to the minimal stable feature.

Given stable features, we encourage the model's predictions always keep invariant under diverse environments.
Specifically, given a group of graphs $\Set{G}=\{g_i\}$, we can obtain the estimated stable and environmental feature representation sets $\Set{H}_{st}$ and $\Set{H}_{en}$ via $\Phi(\cdot)$ and $h(\cdot)$. 
Given a stable feature representation $\Mat{h}_{{st}_i}\in \Set{H}_{st}$, we combine it with all $\Mat{h}_{{en}_j}\in\Set{H}_{en}$ to form a synthesized representation: $\Mat{h}_{(i,j)}=\Mat{h}_{{st}_i}+\Mat{h}_{{en}_j}$.
To ensure invariance, we define the following objective:
\begin{equation}\label{equ:invariance} 
 \mathcal{L}_{sem} = \mathbb{E}_{(g_i, y_i)\sim \mathcal{D}_{tr}}\left\{\frac{1}{|\Set{G}|}\sum_{j=1}^{|\Set{G}|}\ell(\omega(\Mat{h}_{(i,j)}), y_i)\right\}.
\end{equation}
It encourages the independence between labels and environments, conditioned on the given stable features.
Moreover, obtaining ground-truth environments $\Set{E}$ is a challenging task in practice for invariant learning. Consequently, we resort to an unsupervised clustering algorithm, specifically $K$-means \cite{hartigan1979ak}, for inferring the environment.
In particular, we utilize the learned environmental representations $\Set{H}_{en}$ to calculate their distances to the $K$ cluster centers for clustering purposes. Upon the algorithm's convergence, we can segregate the corresponding data into $K$ distinct clusters. Each cluster represents a unique environment, enabling us to derive the environment label $\hat{e}_i\in\hat{\Set{E}}$ for each graph from the estimated environment set $\hat{\Set{E}}$.

Finally, to achieve invariant learning from a unified perspective, we define our final optimization objective:
\begin{equation}\label{equ:final}
\mathcal{L} = \mathcal{L}_{suf} + \alpha\mathcal{L}_{str}  + \beta\mathcal{L}_{sem},
\end{equation}
where $\alpha$ and $\beta$ hyperparameters control the strength of structural and semantic invariant learning.
The overview of the proposed UIL is depicted in Figure \ref{fig:model}.
% Furthermore, for structural invariant learning, the ground-truth environments $\Set{E}$ are difficult to obtain in practice.
% Hence, we adopt an unsupervised clustering algorithm, K-means \cite{hartigan1979ak}, for environment inference.
% Specifically, we use the learned environmental representations $\Set{H}_{en}$ to calculate their distances to $K$ cluster centers for clustering. After the algorithm converges, we can divide the corresponding data into $K$ clusters.
% Each cluster represents an environment, and we can obtain the environment label $\hat{e}_i\in\hat{\Set{E}}$ for each graph from the estimated environment set $\hat{\Set{E}}$.

\section{Theoretical Discussion}\label{sec:the}

In this section, we provide a theoretical analysis to give insights into the proposed optimization objective. For the purpose of the theoretical discussion, we focus solely on the structural information of the graph data, neglecting the node features. We establish our discussion based on an assumption related to the data-generating distribution.

\begin{assumption}
Let $\Set{G} = \{g | g\in\mathbb{G}\}$ be any set of graphs with the same label $y$. We denote its minimal stable feature set and estimated stable features (as per Definition \ref{def:est}) as $\Set{G}_{st}$ and $\hat{\Set{G}}_{st}$, respectively.
We postulate that $\Set{G}_{st}$ and $\hat{\Set{G}}_{st}$ can be generated by graphons $W_{st}$ and $\hat{W}_{st}$, respectively.
\end{assumption}
This assumption is plausible, as a graphon acts as a global feature summarizing the shared and stable structural knowledge across a class of graphs. The structural features of a data subset may not be statically deterministic but could adhere to a certain distribution, such as the Bernoulli distribution. Given that the structures of the minimal stable features demonstrate regular patterns, the values within their respective graphons are likely to approximate 0 or 1. Conversely, stable-environment features, typically irregular within the graph space, would generally see their graphon values uniformly distributed between 0 and 1.

To illustrate the necessity of our proposed structural invariant learning, we consider a toy example. 
For any $G_{st}\in\Set{G}_{st}$ and $\hat{G}_{st}\in\hat{\Set{G}}_{st}$, we define $\Mat{A}_{st}$ and $\hat{\Mat{A}}_{st}$ as the adjacency matrix variables of $G_{st}$ and $\hat{G}_{st}$, respectively. Assume that the stable-environmental feature $G_{se}$ follows a Bernoulli distribution, that is, $\Mat{A}_{se}[i,j]\sim \text{Bernoulli}(p)$, and maintains symmetry (\ie $\Mat{A}_{se}[i,j] = \Mat{A}_{se}[j,i]$) and reflexivity (\ie $\Mat{A}_{se}[i,i] = 0$). Consequently, the estimated stable feature can be expressed as: $\hat{\Mat{A}}_{st} = \Mat{A}_{st} \vee \Mat{A}_{se}$, where $\vee$ denotes an element-wise OR operation between two matrices.

% For any $g_{st}\in\Set{G}_{st}$ and $\hat{g}_{st}\in\hat{\Set{G}}_{st}$, we define $\Mat{A}_{st}$ and $\hat{\Mat{A}}_{st}$ to be the adjacency matrix variable of $g_{st}$ and $\hat{g}_{st}$, respectively.
% Assume that the stable-environmental feature $g_{se}$ obeys the Bernoulli distribution, that is, $\Mat{A}_{se}[i,j]\sim \text{Bernoulli}(p)$, and satisfy symmetry, \ie $\Mat{A}_{se}[i,j] = \Mat{A}_{se}[j,i]$ and reflexivity, \ie $\Mat{A}_{se}[i,i] = 0$.
% Hence, the estimated stable feature can be written as: $\hat{\Mat{A}}_{st} = \Mat{A}_{st} \vee \Mat{A}_{se}$, where $\vee$ symbolizes an element-wise OR operation between two matrices.

\begin{lemma}\label{lemma:add}
Given the graphon of a group of minimal stable features, \ie $W_{st}$, we introduce a Bernoulli distribution with parameter $p$ for the stable-environmental features, whose graphon is denoted by $W_{se}=\mathbf{1}p-\mathbf{I}p$, where $\mathbf{1}$ is an all-one matrix and $\mathbf{I}$ is an identity matrix. Then, the graphon of the estimated stable features is expressed as $\hat{W}_{st} = W_{st} + W_{se} \odot (\mathbf{1}-W_{st})$.
\end{lemma}

% \begin{lemma}\label{lemma:add}
% Given the graphon of a class of the minimal stable features, \ie $W_{st}$, we introduce a Bernoulli adjacency matrix with parameter $p$ for the stable-environmental features, whose graphon is denoted by $W_{se}=\mathbf{1}p-\mathbf{I}p$, where $\mathbf{1}$ and $\mathbf{I}$ is all one matrix and identity matrix. Then the graphon of the estimated stable features is
% $\hat{W}_{st} = W_{st} + W_{se} \odot (\mathbf{1}-W_{st})$.
% \end{lemma}

\begin{proposition}
Given the model $f=\omega \circ h \circ \Phi$, where $\Phi$ is used to extract the estimated stable feature, \ie $\hat{G}_{st}=\Phi(G)$, if the optimization objective is defined as per \Eqref{equ:final}, $\hat{G}_{st}$ will approach the minimal stable feature $G_{st}$ indefinitely.
\end{proposition}
We leave the detailed proofs in Appendix \ref{apd:def_proof}.
Hence, our optimization objective will ultimately converge to the minimal stable features, which theoretically supports our claims.

\begin{table*}[t]
\centering
% \vspace{-2mm}
\caption{Graph classification performance on synthetic and real-world datasets. Numbers in \textbf{bold} indicate the best performance, while the \underline{underlined} numbers indicate the second best performance.}
\vspace{-1mm}
\label{table:main_all}
\resizebox{0.95\textwidth}{!}{\begin{tabular}{clcccccccc}
\toprule
\multirow{2}{*}{Type} & \multirow{2}{*}{Method}  & \multicolumn{2}{c}{Motif} & CMNIST & \multicolumn{2}{c}{Molhiv} & \multicolumn{2}{c}{Molbbbp} & \multirow{2}{*}{Average}  \\ 
\cmidrule(r){3-4}  \cmidrule(r){5-5} \cmidrule(r){6-7} \cmidrule(r){8-9}
& & base & size & color & scaffold & size & scaffold & size  \\
\midrule
\multirow{4}{*}{\makecell[c]{General \\ Generalization}} & ERM         & 81.44\scriptsize{$\pm$2.54} & 70.75\scriptsize{$\pm$2.32} & 28.60\scriptsize{$\pm$1.87} & 72.33\scriptsize{$\pm$1.34} & 63.26\scriptsize{$\pm$1.98} & 68.10\scriptsize{$\pm$1.68} & 78.29\scriptsize{$\pm$3.76} & 66.11 \\
& IRM         & 80.71\scriptsize{$\pm$2.81} & 69.77\scriptsize{$\pm$3.88} & 27.83\scriptsize{$\pm$2.13} & 72.59\scriptsize{$\pm$1.87} & 59.90\scriptsize{$\pm$1.60} & 67.22\scriptsize{$\pm$1.15} & 77.56\scriptsize{$\pm$2.48} & 65.08 \\
& VREx      & 81.56\scriptsize{$\pm$2.14}  & 70.24\scriptsize{$\pm$4.90} & 28.48\scriptsize{$\pm$2.87} & 72.60\scriptsize{$\pm$2.03} & 60.23\scriptsize{$\pm$1.55} & 68.74\scriptsize{$\pm$1.03} & 78.76\scriptsize{$\pm$2.37} & 65.80 \\
& GroupDRO & 81.43\scriptsize{$\pm$2.69} & 69.98\scriptsize{$\pm$4.73} & 29.07\scriptsize{$\pm$3.14} & \underline{73.64\scriptsize{$\pm$2.57}} & 61.37\scriptsize{$\pm$2.82} & 66.47\scriptsize{$\pm$2.39} & 79.27\scriptsize{$\pm$2.43} & 65.89 \\
\midrule
\multirow{5}{*}{\makecell[c]{Graph \\ Augmentation}} & DropEdge   & 78.97\scriptsize{$\pm$3.41} & 55.27\scriptsize{$\pm$5.93} & 22.65\scriptsize{$\pm$2.90} & 66.78\scriptsize{$\pm$2.68} & 54.92\scriptsize{$\pm$1.73} & 66.49\scriptsize{$\pm$1.55} & 78.32\scriptsize{$\pm$3.44} & 60.49 \\
& FLAG  & 80.91\scriptsize{$\pm$1.04} & 56.26\scriptsize{$\pm$3.98} & \underline{32.30\scriptsize{$\pm$2.69}} & 70.45\scriptsize{$\pm$1.55} & 66.44\scriptsize{$\pm$2.32} & 67.69\scriptsize{$\pm$2.36} & 79.26\scriptsize{$\pm$2.26} & 64.76 \\
& M-Mixup   & 77.63\scriptsize{$\pm$2.99} & 67.81\scriptsize{$\pm$3.32} & 26.47\scriptsize{$\pm$3.45} & 72.03\scriptsize{$\pm$1.25} & 64.87\scriptsize{$\pm$1.88} & 68.75\scriptsize{$\pm$0.34} & 78.92\scriptsize{$\pm$2.43} & 65.21 \\
& $\Set{G}$-Mixup  & 74.66\scriptsize{$\pm$1.89} & 59.92\scriptsize{$\pm$2.10} & 31.85\scriptsize{$\pm$5.82} & 71.69\scriptsize{$\pm$1.74} & 70.53\scriptsize{$\pm$2.02} & 67.44\scriptsize{$\pm$1.62} & 78.55\scriptsize{$\pm$4.16} & 64.95 \\
& GREA  & 80.60\scriptsize{$\pm$2.49} & \underline{73.31\scriptsize{$\pm$1.85}} & 29.02\scriptsize{$\pm$3.26} & 70.96\scriptsize{$\pm$3.16} & 66.48\scriptsize{$\pm$4.13} &  \underline{69.72\scriptsize{$\pm$1.66}} & 77.34\scriptsize{$\pm$3.52} & \underline{66.78}  \\
\midrule
\multirow{8}{*}{\makecell[c]{Graph \\ Generalization}} & DIR     & 72.14\scriptsize{$\pm$6.37}  & 56.28\scriptsize{$\pm$7.83} & 33.20\scriptsize{$\pm$6.17} & 69.05\scriptsize{$\pm$2.60} & \underline{72.61\scriptsize{$\pm$2.31}} & 66.86\scriptsize{$\pm$2.25} & 76.40\scriptsize{$\pm$4.43} & 63.79 \\
& CAL       & 81.94\scriptsize{$\pm$1.20} & 71.44\scriptsize{$\pm$2.86} & 27.99\scriptsize{$\pm$3.24} & 70.15\scriptsize{$\pm$2.14} & 62.36\scriptsize{$\pm$1.42} & 68.06\scriptsize{$\pm$2.60} & \underline{79.50\scriptsize{$\pm$4.81}} & 65.92 \\
& GSAT     & \underline{83.71\scriptsize{$\pm$2.30}} & 64.16\scriptsize{$\pm$3.35} & 28.17\scriptsize{$\pm$1.26} & 68.88\scriptsize{$\pm$1.96} & 65.63\scriptsize{$\pm$0.88} & 66.78\scriptsize{$\pm$1.45} & 75.63\scriptsize{$\pm$3.83} & 64.71 \\
& OOD-GNN  & 80.22\scriptsize{$\pm$2.28} & 68.62\scriptsize{$\pm$2.98}  & 26.49\scriptsize{$\pm$2.94} & 70.45\scriptsize{$\pm$2.02} & 57.49\scriptsize{$\pm$1.08} & 66.72\scriptsize{$\pm$1.23} & 79.48\scriptsize{$\pm$4.19} & 64.21 \\
& StableGNN  & 73.04\scriptsize{$\pm$2.78} & 59.83\scriptsize{$\pm$3.40} & 28.38\scriptsize{$\pm$3.49} & 68.23\scriptsize{$\pm$2.44} & 58.33\scriptsize{$\pm$4.69} & 66.74\scriptsize{$\pm$1.30} & 77.47\scriptsize{$\pm$4.69} & 61.72 \\
& CIGA  & 75.01\scriptsize{$\pm$3.56} & 70.38\scriptsize{$\pm$6.61} & 32.22\scriptsize{$\pm$2.67} & 72.88\scriptsize{$\pm$2.89} & 70.32\scriptsize{$\pm$2.60} & 64.92\scriptsize{$\pm$2.09} & 65.98\scriptsize{$\pm$3.31} & 64.53 \\
& DisC  & 76.70\scriptsize{$\pm$0.47} & 53.34\scriptsize{$\pm$13.71} & 24.99\scriptsize{$\pm$1.78} & 69.35\scriptsize{$\pm$3.11} & 64.96\scriptsize{$\pm$5.23} & 67.12\scriptsize{$\pm$2.11} & 56.59\scriptsize{$\pm$10.09} & 59.01 \\
& UIL (ours) & \textbf{86.97\scriptsize{$\pm$1.67}} & \textbf{74.80\scriptsize{$\pm$4.81}} & \textbf{35.68\scriptsize{$\pm$2.43}} & 
\textbf{73.76\scriptsize{$\pm$2.29}} & \textbf{74.33\scriptsize{$\pm$1.32}} & \textbf{70.52\scriptsize{$\pm$0.95}} & \textbf{80.29\scriptsize{$\pm$2.84}} & \textbf{70.90} \\ 
\bottomrule
\end{tabular}}
\vspace{-2mm}
\end{table*}

\section{Experiments}

In this section, we undertake comprehensive experiments on synthetic and real-world datasets to assess the efficacy of the proposed frameworks relative to state-of-the-art methods. Specifically, we endeavor to address the following research questions:
\begin{itemize}[leftmargin=4mm]
% \item \textbf{RQ1:} To what extent does our proposed UIL mitigate the OOD issue in graph classification?
\item \textbf{RQ1:} How does the performance of UIL compare to existing state-of-the-art methods?
\item \textbf{RQ2:} Does UIL excel at learning minimal stable features more proficiently than the current methodologies?
\item \textbf{RQ3:} Is UIL adept at uncovering the structural invariant knowledge veiled within a group of graphs?
% \item \textbf{RQ4:} Is UIL more adept at uncovering the structural knowledge veiled within a class of graph data than its current counterparts?
% \item \textbf{RQ5:} How do the different components and hyperparameters of UIL influence the overall performance?
\end{itemize}
% We adopted SYN \cite{sui2021causal}, GOOD \cite{gui2022good} and OGB datasets \cite{hu2020open}, which are benchmark datasets for graph OOD problems.
% Experimental settings, such as datasets, metrics, baselines, and hyperparameters, are provided in Appendix \ref{apd:implementation}.
% \begin{itemize}[leftmargin=4mm]
% \item \textbf{RQ1:} How effective is our proposed MGCL for alleviating the OOD issue in graph classification?
% \item \textbf{RQ2:} How does MGCL perform compared to state-of-the-art graph OOD generalization methods?
% \item \textbf{RQ3:} What causal features or generative patterns have MGCL learned from local and global perspectives?
% \item \textbf{RQ4:} How do the different components and hyperparameters of MGCL affect performance?
% \end{itemize}
We put the detailed settings of the experiment, such as datasets, baseline methods and training hyperparameters in Appendix \ref{apd:details}.

\subsection{Comparison with State-of-the-Arts (RQ1)}

To demonstrate the superiority of UIL, we compare with numerous state-of-the-art efforts, including general generalization algorithms, graph generalization methods and augmentation methods.
We choose GOOD \cite{gui2022good} and OGB \cite{hu2020open} datasets, including Motif, CMNIST, Molhiv and Molbbbp, and create distribution shifts with diverse graph features, such as base, color, scaffold and size.
For a fair comparison, we adopt the GIN model to conduct all experiments. 
From the results in Table \ref{table:main_all}, the direct application of general generalization algorithms to graph learning tasks does not effectively enhance generalization. On average, these methods yield performances ranging from 65.08 to 65.89, which are comparable to the performance of ERM, \ie 66.11. Secondly, existing graph generalization algorithms do not consistently outperform ERM. Specifically, algorithms such as OOD-GNN and StableGNN often fail to surpass ERM. While invariant learning methods like DIR, CAL, and DisC do outperform ERM in certain instances, such as DIR improving ERM by 14.78\% on Molhiv (size) and GSAT improving ERM by 2.78\% on Motif (base), they also underperform in other scenarios. For instance, compared to ERM, DIR's performance drops by 20.45\% in Motif (size), and GSAT's performance decreases by 3.39\% in Molbbbp (size).
In comparison to existing methods, UIL consistently enhances the OOD generalization. Firstly, substantial performance improvements are observed with UIL on synthetic datasets. Specifically, for the Motif dataset, UIL increases accuracy by 6.26\% and 5.72\% compared to ERM across two domains (\ie base and size). Compared with the best performing baseline methods, GSAT and GREA, the improvement is 3.38\% and 2.03\%, respectively. For CMNIST, UIL exhibits an improvement of 24.75\% and 10.46\% over ERM and the top-performing baseline, DIR, respectively.
These results underscore the importance of ensuring both structural and semantic invariance principles.

\begin{figure}[t] 
\centering
\subfigbottomskip=2mm
\subfigcapskip=-1mm
\subfigure[Precision ($b=0.3$)]{
\label{fig:equ_inv.dir.equ}
\includegraphics[width=0.48\linewidth]{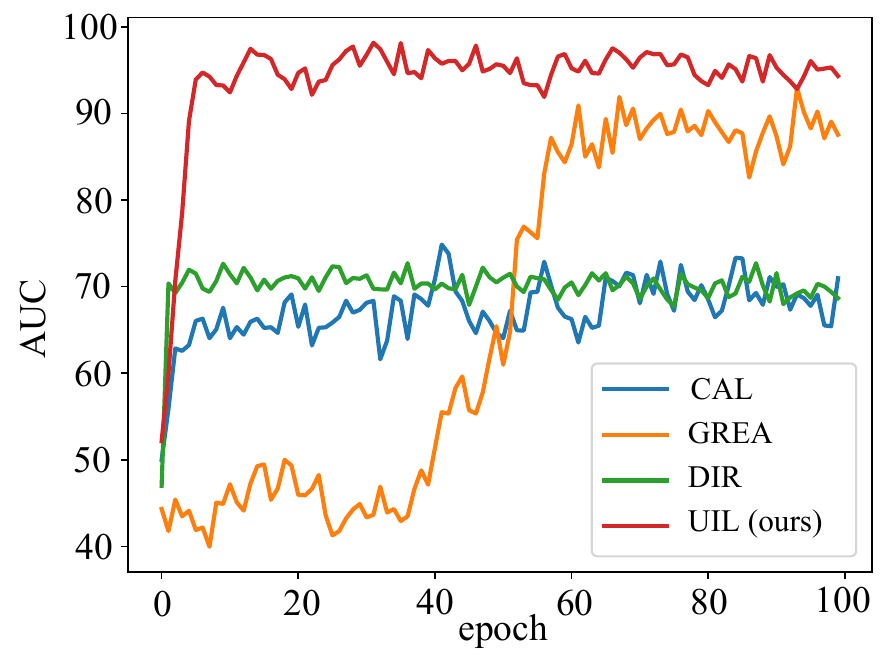}}
\hspace{-2mm}
\subfigure[Accuracy ($b=0.3$)]{
\label{fig:equ_inv.grea.equ}
\includegraphics[width=0.48\linewidth]{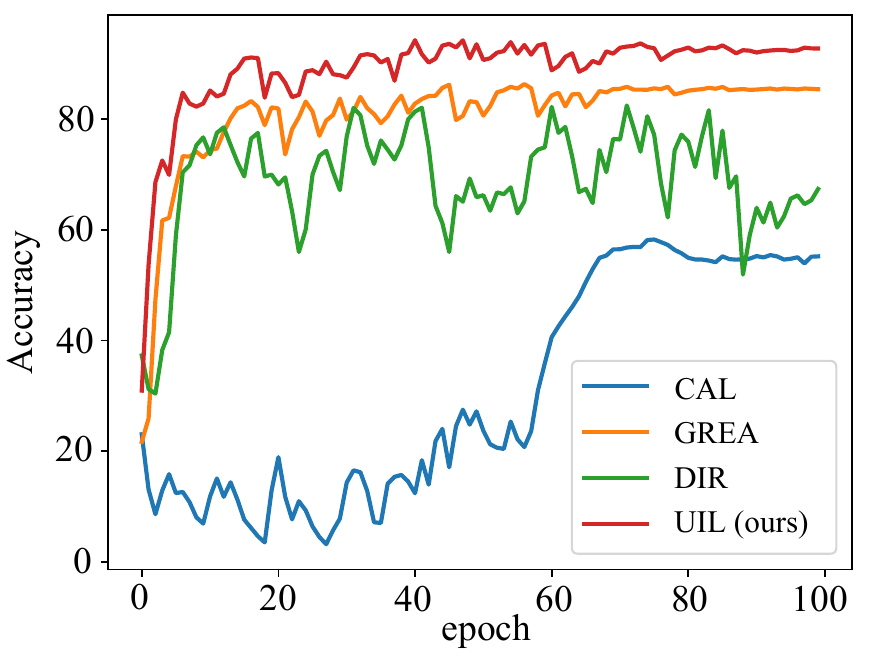}}
\subfigure[Precision ($b=0.7$)]{
\label{fig:equ_inv.cal.equ}
\includegraphics[width=0.48\linewidth]{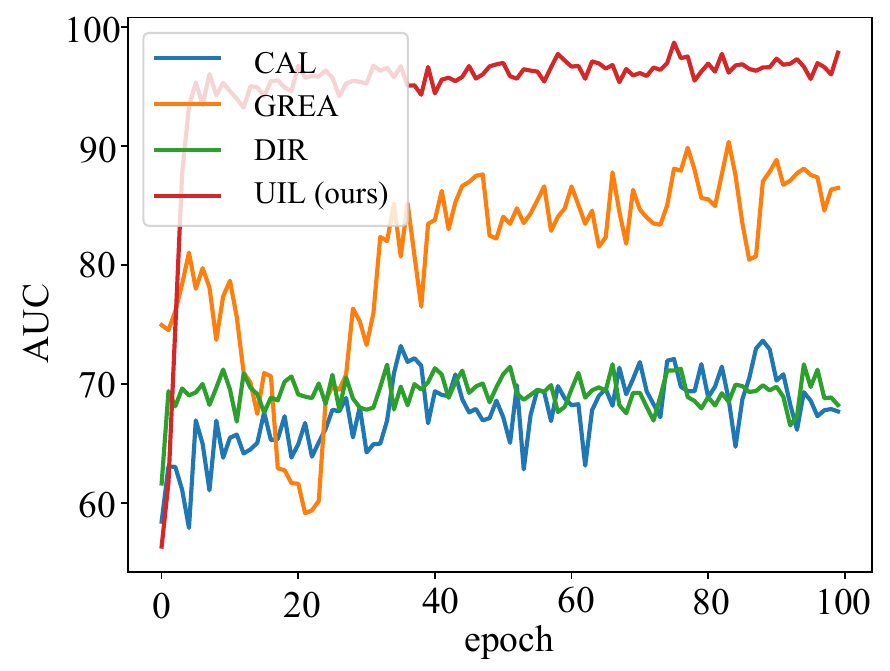}}
\hspace{-2mm}
\subfigure[Accuracy ($b=0.7$)]{
\label{fig:equ_inv.our.equ}
\includegraphics[width=0.48\linewidth]{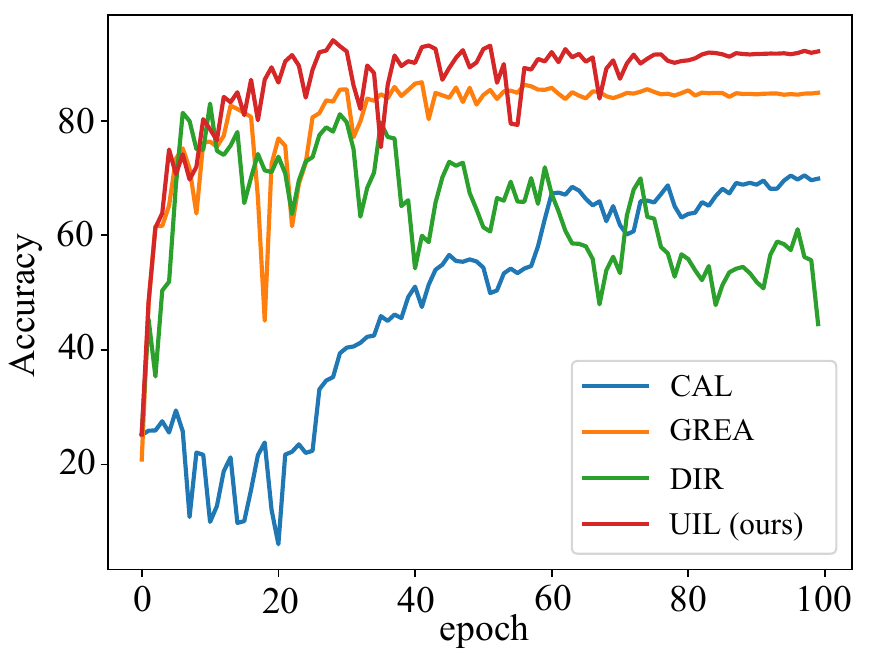}}
\vspace{-2mm}
\caption{Precision of extracting the minimal stable features (AUC) and  classification accuracy (\%).}
\label{fig:auc}
\vspace{-4mm}
\end{figure}

\begin{figure*}[t]
\centering
\includegraphics[width=0.95\linewidth]{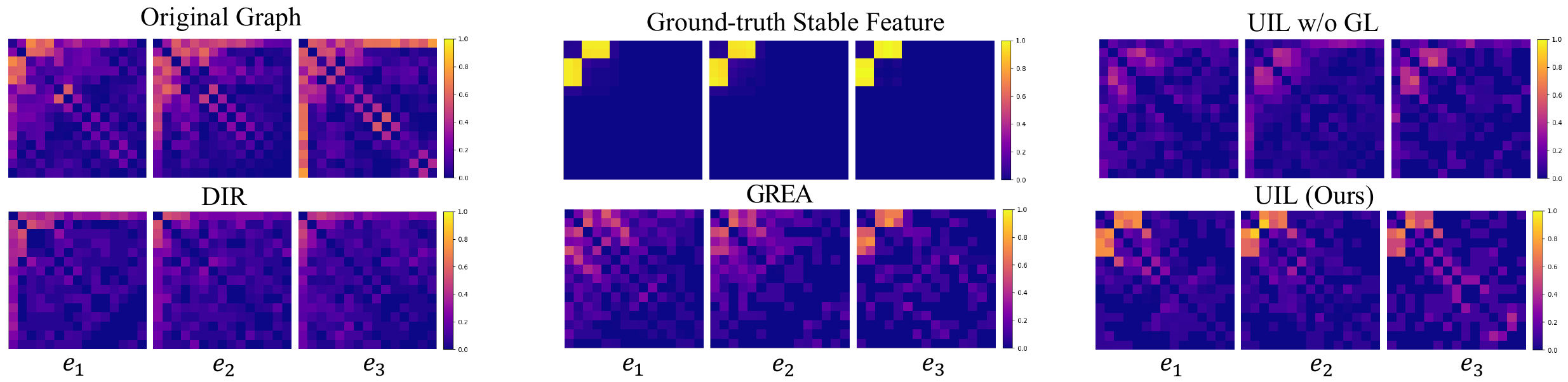}
\vspace{-2mm}
\caption{Visualization results of learned stable graphons. $e_1, e_2, e_3$ refer to different environments in Motif dataset. }
\label{fig:vis_global}
\vspace{-2mm}
\end{figure*}

% \vspace{-4mm}
\subsection{Learning Minimal Stable Features (RQ2)}\label{sec:minimal-stable-feature}

We claim that UIL is adept at achieving structural invariance, thus enabling it to efficiently identify minimal stable features. 
In this section, we present experiments designed to provide experimental validation for our assertions and contrast UIL's performance with state-of-the-art methodologies.

\textbf{Capturing Superior Stable Features.}
We design extensive experiments to affirm that our structural invariance principle accurately captures the minimal stable features, while concurrently eliminating other environmental features. Specifically, we utilize the toy example referenced in our theoretical discussion (refer Section \ref{sec:the}), and subsequently construct a novel dataset derived from the SYN-$b$ dataset.
Our dataset is supplemented with Bernoulli subgraphs. These subgraphs, characterized by their invariance with alterations in training environments, serve as a representation of additional stable-environmental features. Throughout the training phase, we curate a dataset by setting parameter $p=0.001$. During the testing phase, we modify this parameter to $p=0.003$.
We evaluate both the classification accuracy and the precision of extracting the minimal stable subgraph, \ie comparing $\hat{G}_{st}$ with $G_{st}$. In keeping with studies \cite{ying2019gnnexplainer,luo2020parameterized}, we employ the interpretation evaluation metric ROC AUC to quantify precision. Intuitively, this metric signifies the extent of overlap between the estimated stable features $\hat{G}_{st}$ and minimal stable features $G_{st}$.
We conduct experiments under two distinct settings, $b=0.3$ and $b=0.7$, and compare our results with methods predominantly based on semantic invariance. The outcomes of these experiments are depicted in Figure \ref{fig:auc}, and our principal observations are as follows:

In terms of the precision of capturing the minimal stable features, we discern that methods exclusively reliant on semantic invariance encounter difficulties in accurately deducing $\hat{G}_{st}\rightarrow G_{st}$. As expounded in our theoretical discussion, the presence of stable-environmental features likely contributes to this. Additionally, methodologies like DIR and GREA, which employ a fixed stable ratio, struggle to converge to $G_{st}$.
Contrastingly, UIL, considering both structural invariance and a dynamic stable ratio, is proficient at accurately identifying $G_{st}$.
As anticipated, UIL consistently outperforms all baseline methods in terms of classification accuracy. This substantiates our assertion that exclusive reliance on semantic invariance is insufficient for ensuring generalization across test distributions. UIL, considering both semantic and structural invariance, effectively attains generalization.
These empirical results provide further evidence supporting the validity of our theoretical propositions.

\begin{table}[t]
\centering
\vspace{-2mm}
\caption{\centering Cut distance comparisons.}
\vspace{-4mm}
\label{table:cut}
\resizebox{0.45\textwidth}{!}{\begin{tabular}{lcccc}
\toprule
Method & ${\rm Dis}(e_1, e_2)$  & ${\rm Dis}(e_2, e_3)$  & ${\rm Dis}(e_1, e_3)$ & Avg.  \\
\toprule
% \hline
Original graph   & 11.09\scriptsize{$\pm$1.36}  & 11.91\scriptsize{$\pm$1.41}  & 14.10\scriptsize{$\pm$1.69} & 12.37 \\
Sta-subgraph (GT) & 2.49\scriptsize{$\pm$3.49} & 2.26\scriptsize{$\pm$3.24} & 1.85\scriptsize{$\pm$2.40} & 2.20 \\
Env-subgraph (GT) & 9.72\scriptsize{$\pm$2.05} & 10.88\scriptsize{$\pm$1.31} & 13.91\scriptsize{$\pm$1.68} & 11.50 \\
\midrule
DIR         & 7.92\scriptsize{$\pm$1.15} & 7.87\scriptsize{$\pm$1.38} & 7.92\scriptsize{$\pm$1.41} & 7.90  \\
GREA        & 6.25\scriptsize{$\pm$1.75}  & 6.05\scriptsize{$\pm$1.39} & 5.86\scriptsize{$\pm$1.25} & 6.05  \\
CAL         & 6.94\scriptsize{$\pm$1.14}  & 7.57\scriptsize{$\pm$1.11} & 8.83\scriptsize{$\pm$1.39} & 7.78 \\
UIL w/o GL & 7.24\scriptsize{$\pm$2.41}  & 7.00\scriptsize{$\pm$2.68} & 7.58\scriptsize{$\pm$1.74} &  7.27 \\
UIL (ours) & \textbf{4.40\scriptsize{$\pm$2.35}}  & \textbf{4.24\scriptsize{$\pm$2.49}}  & \textbf{5.32\scriptsize{$\pm$2.07}}  & \textbf{4.65}\\
\bottomrule
\end{tabular}}
\vspace{-4mm}
\end{table}

\subsection{Evaluation of Structural Invariance (RQ3)}\label{sec:exp:str}

\textbf{Cut Distance Comparisons.}
We randomly select a group of graphs with the label $y$ and the environment $e_i$ from the dataset, and we can obtain the learned stable graphons $W_{e_i}^{y}$.
And we define our distance metric as: 
\begin{equation}
{\rm Dis}(e_i, e_j)=\frac{1}{C}\sum_{y=1}^C\delta_{\Box}(W_{e_i}^{y},W_{e_j}^{y}).
\end{equation}
To calculate the cut distance $\delta_{\Box}(\cdot,\cdot)$, we employ the GWB algorithm \cite{xu2021learning}. We use the stable features identified by these methods to estimate graphons. We also consider a variant of UIL, UIL w/o GL, which omits the graphon learning component, \ie \Eqref{equ:graphon_loss2}. 
The experimental results are presented in Table \ref{table:cut}. The terms "Original graph", "Sta-subgraph (GT)", and "Env-subgraph (GT)" refer to the direct calculation of the distance using the original full graphs, ground-truth stable subgraphs, and environmental subgraphs, respectively. 
From the results of the first three rows, it is clear that graphons generated by original graphs and environmental subgraphs maintain substantial distances. In contrast, stable subgraphs exhibit smaller distances. This substantiates the notion that graphs with the same label possess global structural patterns or class-specific knowledge that remains invariant across varying environments.
UIL consistently exhibits smaller distances compared to the baselines, demonstrating that UIL is indeed capable of adhering to the structural invariance principle to capture stable graphons effectively. When graphon learning is omitted, the distance increases significantly, reinforcing the importance of structural invariant learning.

\begin{figure}[t]
\centering
\includegraphics[width=0.9\linewidth]{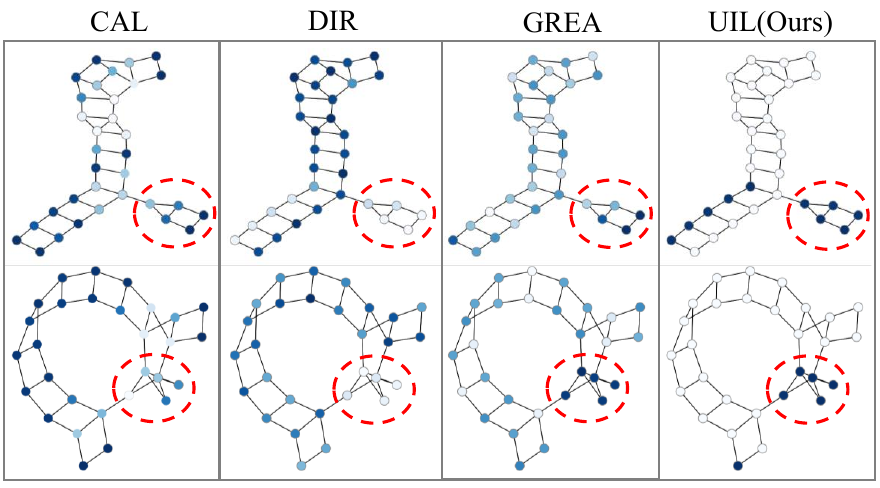}
\vspace{-2mm}
\caption{Visualization results of the captured stable features.}
\label{fig:vis_local}
\vspace{-6mm}
\end{figure}

\textbf{Visualizations.}
We visualize the captured stable graphons in Figure \ref{fig:vis_global}.
``w/o GL'' denotes UIL without graphon learning.
We can find that our method can be closer to the ground-truth stable graphon.
To provide an intuitive comparison between the estimated stable subgraphs $\hat{G}_{st}$ generated by our method and those of baseline methodologies, we also conduct visualization experiments in Figure \ref{fig:vis_local}. Areas of darker shading denote larger stable mask values, and the ground-truth stable features are highlighted using red circles. As can be observed, UIL more accurately captures the stable features, while adeptly disregarding other environmental features.

\subsection{Ablation Study}

\begin{figure}[t]
\vspace{-2mm}
\centering
\vspace{-2mm}
\subfigure[Different components of UIL]{
\label{fig:ablation_hist_loss.1}
\begin{minipage}[t]{0.22\textwidth}
\centering
\includegraphics[width=1\textwidth]{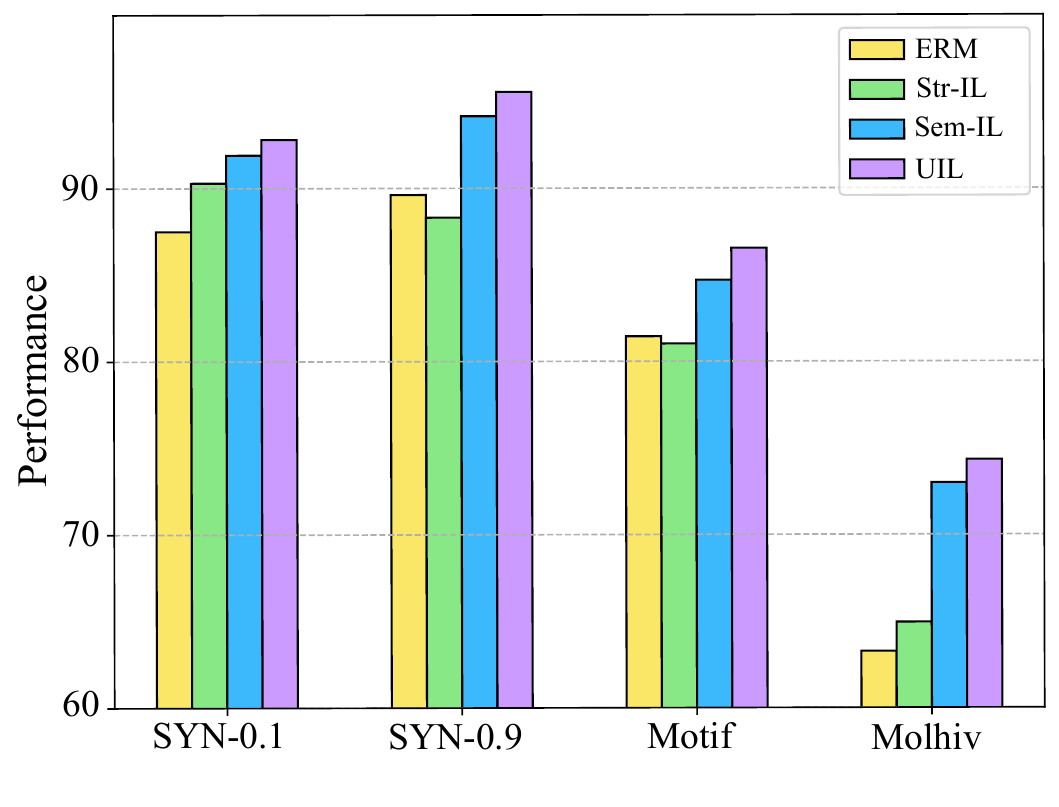}
\end{minipage}%
}%
\hspace{1mm}
\subfigure[Loss coefficients: $\alpha$ and $\beta$]{
\label{fig:ablation_hist_loss.2}
\begin{minipage}[t]{0.22\textwidth}
\centering
\includegraphics[width=1\textwidth]{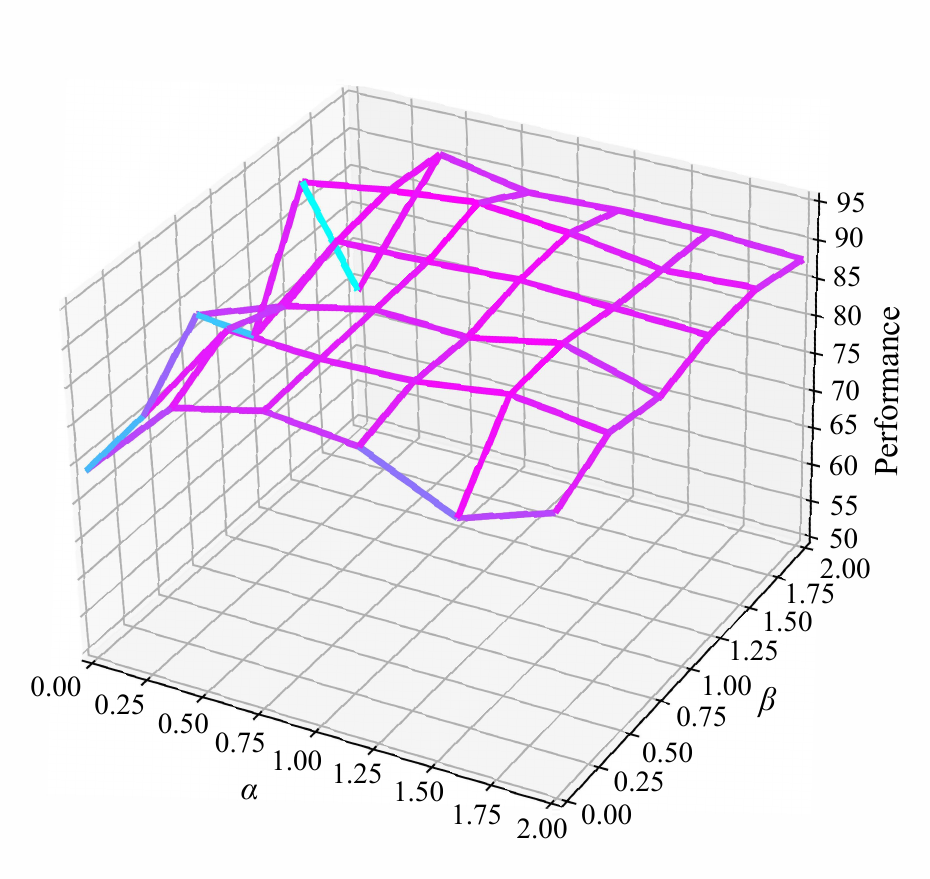}
\end{minipage}%
}%
\centering
\vspace{-2mm}
\caption{Classification performance over different components of UIL and loss coefficients $\alpha$ and $\beta$. 
}
\vspace{-2mm}
\label{fig:ablation_hist_loss}
\end{figure}

\textbf{Different Components in UIL.} 
We explore the impact of semantic and structural invariance.
The results are shown in Figure \ref{fig:ablation_hist_loss} (a), where ``Sem-IL'' means UIL without graphon learning; ``Str-IL'' means directly using whole features for graphon learning without semantic invariance.
We can find that it is difficult to achieve better results for semantic invariance or structural invariance alone, and sometimes even worse than ERM.
While invariant learning from both semantic and structural views can achieve the best results.

\begin{figure}[t]
\vspace{-2mm}
\centering
\subfigcapskip=-5pt
\subfigure[Initial stable ratio]{
\label{fig:ablation_causal_env.1}
\begin{minipage}[t]{0.22\textwidth}
\centering
\includegraphics[width=1\textwidth]{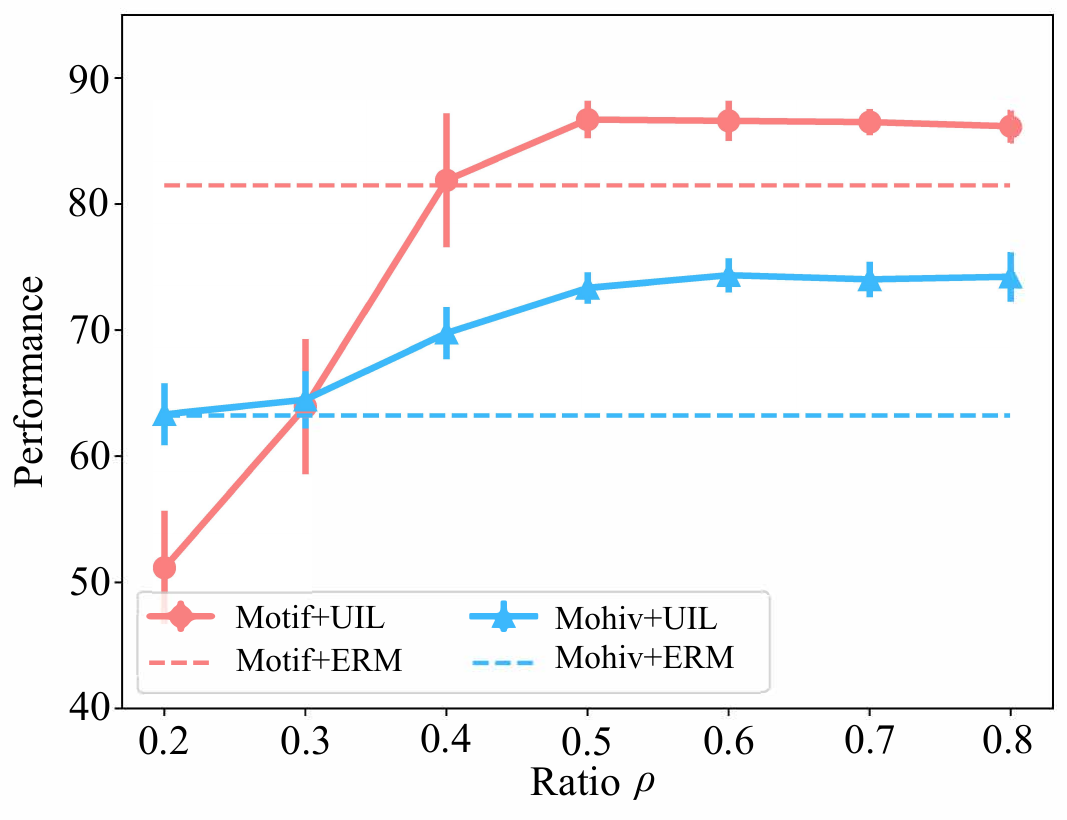}
\end{minipage}%
}%
\hspace{1mm}
\subfigure[Environments]{
\label{fig:ablation_causal_env.2}
\begin{minipage}[t]{0.22\textwidth}
\centering
\includegraphics[width=1\textwidth]{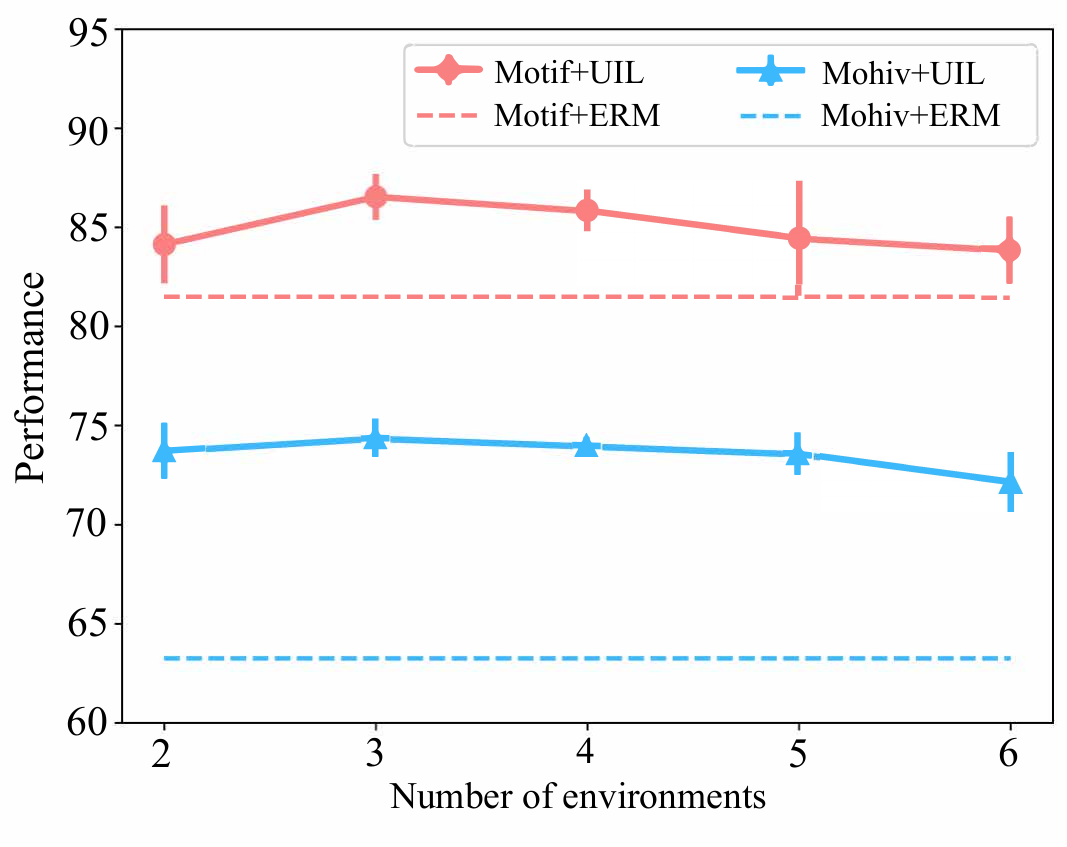}
\end{minipage}%
}%
\centering
\vspace{-4mm}
\caption{Classification performance over $\rho$ and $|\hat{\Set{E}}|$.}
\vspace{-4mm}
\label{fig:ablation_causal_env}
\end{figure}

\vspace{2mm}
\noindent\textbf{Hyperparameter Sensitivity.}
Firstly, we explore the sensitivities of $\alpha$ and $\beta$ in Figure \ref{fig:ablation_hist_loss} (b).
We find that the performance is insensitive to $\alpha$ or $\beta$.
Secondly, Figure \ref{fig:ablation_causal_env} (a) shows that smaller or larger initial values of stable ratio $\rho$ will reduce the performance, and the best value is around 0.5$\sim$0.7.
Thirdly, Figure \ref{fig:ablation_causal_env} (b) shows that the performance is also insensitive to the number of inferred environments $|\hat{\Set{E}}|$.

\section{Related Work}

\textbf{Out-of-Distribution Generalization on Graphs.}
Graph neural networks (GNNs), despite their significant success in numerous graph-related tasks, are often plagued by issues concerning interpretability \cite{fang2024exgc}, robustness \cite{jin2020graph,sui2022towards,gao2024graph}, and efficiency \cite{chen2021unified,JCST-2206-12583,wang2022exploring,wang2024dynamic,zhangtwo,wu2023gif}. 
Current research \cite{gui2022good,li2022out} indicates the presence of the OOD issue in various graph learning tasks, including graph classification \cite{DIR,liu2022graph} and node classification \cite{wu2022handling,liu2023flood}. 
% OOD problems in graph data can stem from several factors, such as graph size \cite{knyazev2019understanding,liu2022size,bevilacqua2021size,yehudai2021local,buffellisizeshiftreg}, structural shift \cite{gao2023alleviating}, concept shift \cite{DIR,sui2021causal}, covariate shift \cite{sui2022adversarial}, or label shift \cite{yu2023mind,zhu2021shift}. 
This has led to an increasing interest in OOD generalization in graph-related tasks, spawning methodologies based on graph data augmentation \cite{wang2021mixup,kong2022robust,han2022g,luo2024masked,sui2024unleashing,sui2024simple}, stable learning \cite{li2021ood,fan2021generalizing}, and invariant learning \cite{DIR,wu2022handling,sui2021causal,sui2024enhancing,sui2024invariant,li2022learning,wang2024unleashing}.
In this context, invariant learning has emerged as a significant line of research for OOD generalization. It primarily assumes that graph data contain stable features that have a causal relationship with the label, a relationship that persists across different environments. To capture these stable features, various techniques are employed. For example, DIR \cite{DIR} encourages model predictions to remain invariant by intervening on environmental features, while GREA \cite{liu2022graph}, CAL \cite{sui2021causal}, and DisC \cite{fan2022debiasing} motivate the model to make predictions based on invariant stable features through the random combination or replacement of environmental features. 
Furthermore, GALA \cite{chen2024does} establishes key assumptions essential for graph invariant learning.
% However, these methods tend to prioritize learning semantic invariant features, often overlooking structural invariance. 
Consequently, there is a crucial need for graph learning methods that consider both structural and semantic invariance in a comprehensive manner, an area that remains largely unexplored.

\noindent\textbf{Out-of-Distribution Generalization.}
Performance degradation is common when training and test data follow disparate distributions—an issue known as out-of-distribution (OOD) generalization \cite{shen2021towards,ye2022ood}. To alleviate this problem, recent studies have proposed several strategies. For instance, GroupDRO \cite{sagawa2019distributionally} minimizes risk within groups that perform the poorest. IRM \cite{arjovsky2019invariant} operates under the assumption that there are environment-invariant features present in the data and encourages models to capture these features. Other studies, such as JTT \cite{liu2021just} and EIIL \cite{creager2021environment}, utilize a two-stage training framework and environment inference respectively, to enhance the performance of underperforming groups. A range of other methodologies based on causality \cite{wang2021causal}, invariant learning \cite{chang2020invariant}, stable learning \cite{shen2020stable}, information bottlenecks \cite{ahuja2021invariance}, data augmentation \cite{zhang2017mixup}, and test time adaptation \cite{wang2020tent} have also been proposed to improve model generalization. However, due to the inherent irregularities in graph structures, the direct application of these methods to graph learning tasks can prove challenging and often result in suboptimal performance.

\section{Conclusion}

In this paper, we introduce an invariant learning framework, UIL, adept at unearthing superior stable features and significantly mitigating the OOD problem in graph classification tasks. 
UIL offers a unified view for invariant learning, which concurrently accounts for both structural and semantic invariance, to harness superior stable features. 
Specifically, in the graph space, UIL maintains structural invariance by identifying a common generative pattern in a class of stable features via the invariant learning of stable graphons. We employ the cut distance to measure the discrepancies between graphons, with an aspiration that the learned stable graphons will remain invariant across various environments, hence fortifying structural invariance.
Simultaneously, within the semantic space, UIL ensures that the identified stable features maintain consistent and optimal performance across diverse environments, thereby preserving semantic invariance.
Theoretical and empirical proofs substantiate the effectiveness of two invariance principles, highlighting UIL's enhanced ability to discover better stable features. 

% \section*{Acknowledgments}

\begin{acks}
This research is supported by the National Natural Science Foundation of China (92270114), Ant Group through CCF-Ant Research Fund and CCF- AFSG Research Fund. This work is also supported by the University Synergy Innovation Program of Anhui Province (GXXT-2022-040).
\end{acks}

\bibliographystyle{ACM-Reference-Format}
\bibliography{main}
\appendix
\balance

\clearpage

\section{Definitions and Proofs}\label{apd:def_proof}

\subsection{Graphon}

% To address the limitations imposed by semantic invariance, it is necessary to incorporate additional invariance constraints directly within the graph space. In most case, the minimal stable features within the same class exhibit strong structural consistency. For instance, in the MNIST superpixel graph, similar handwritten digit patterns are shared across diverse environments within each class. Similarly, for molecular graphs, a stable feature like the ``-COOH'' functional group determines the acidity. Given a set of acidic molecules distributed across various environments, most will share common patterns based on the ``-COOH'' group.
% Hence, these patterns, serving as the embodiment of class-specific knowledge, implicitly pervade each class of graphs. 
% This suggests that we can impose structural constraints on a set of subgraphs with identical labels in the subgraph space, aspiring for a structure that remains invariant under different environments. However, implementing invariance constraints directly within the graph space is challenging due to the non-Euclidean and irregular structures of graph data, such as varying numbers of nodes and edges.
% Fortunately, inspiration can be drawn from graphon \cite{lovasz2012large}, which can serve as graph functions capable of efficiently capturing shared generative patterns within a set of graph data. The use of graphons has shown potential for overcoming these challenges and aiding in the imposition of invariance constraints within the graph space.

\begin{definition}[Graphon]\label{def:graphon}
A graphon is a symmetric, two-dimensional, and continuous measurable function $W:\Omega^2\rightarrow [0,1]$, where $\Omega$ is a measurable space, \eg $\Omega=[0,1]$. Here symmetric means $W(x,y)=W(y,x)$ for all $x,y\in \Omega$. 
\end{definition}

Given a certain generation rule of graphs, graphon is considered to continuously add nodes according to this rule until infinity, and finally obtain a probability density function. It describes the probability that an edge exists between two nodes. Specifically, given two nodes $v_i, v_j \in [0,1]$, $W(i,j)$ describes the probability that an edge exists between these nodes. 
Indeed, to measure the distance between graphons, it's essential to introduce a suitable distance function.
We first begin with the cut norm \cite{lovasz2012large}.
\begin{definition}[Cut Norm]
    The cut norm of graphon $W$ is defined by
    $$||W||_{\Box}=\sup_{\Set{S},\Set{T}\subseteq \Omega}\left|\int_{\Set{S}\times \Set{T}}W(x,y)dxdy\right|,$$
    where the supremum is taken over all subsets $\Set{S}$ and $\Set{T}$. 
    % \zeminC{上面Equation中S和T换成斜体？}
\end{definition}
The cut norm is the integral of the graphon in the optimal rectangle, and the suitable distance between any two graphons is defined by the cut norm.
Below we measure the distance between different graphons by defining cut distance \cite{lovasz2012large}.
\begin{definition}[Cut Distance]\label{def:cut}
    Given two graphons $W_1$ and $W_2$: $\Omega^2\rightarrow [0,1]$, we define their cut distance as:
    $$\begin{aligned}
    \delta_{\Box}(W_1,W_2)&:=\inf_{\phi}||W_1-W_2^\phi||_{\Box},
    \end{aligned}$$
    where $||\cdot||_{\Box}$ denotes the cut norm, and the infimum is taken over all invertible measure-preserving maps $\phi: \Omega\rightarrow\Omega$, whose absolute value of the determinant of the Jacobian matrix is 1.
\end{definition}
% The symbol $||\cdot||_{\Box}$ denotes the cut norm \cite{lovasz2012large}.
The cut distance offers a methodology for quantifying the difference between two graphons. Initially, it measures the maximum difference of the integral of the two graphons over measurable boxes of $\Omega^2$. Subsequently, it minimizes this maximum difference over all possible measure-preserving maps. As such, the cut distance serves as a tool for comparing two graphons and can facilitate the direct implementation of invariance constraints within the graph data space. Utilizing the cut distance in this way not only provides a metric for graphon comparison but also advances the application of invariance principles within the inherently complex and irregular realm of graph data.

\subsection{Proofs}\label{apd:proof}
We present a brief proof of the proposition as follows. The estimated stable graphon is defined as $\hat{W}_{st} = W_{st} + W_{se} \odot (\mathbf{1}-W_{st})$. By optimizing $\Set{L}_{suf}$ and $\Set{L}_{sem}$, we essentially adhere to the principle of semantic invariance to extract stable features, thereby preventing the feature extractor from capturing features from unstable environments. Additionally, an error exists between the graphon we estimate in each sampling process and the real graphon, \ie $||\hat{W}-W||_2>\mu$ for some $\mu>0$. Thus, we can model the estimated graphon as a Gaussian distribution $\hat{W}_{st}[i,j]\sim \mathcal{N}(W_{st}[i,j]+p(1-W_{st}[i,j]), \epsilon p)$.
In our structural invariance objective as defined in \Eqref{equ:graphon_loss2}, we minimize $||\hat{W}_{st}^{e}-\hat{W}_{st}^{e'}||_2$. Considering
$$
\hat{W}_{st}^{e}[i,j]-\hat{W}_{st}^{e'}[i,j]\sim \mathcal{N}(0, \epsilon(p+p)),
$$
the objective corresponds to minimizing a Gaussian matrix with a mean of $0$ and a variance of $\epsilon(p+p)$. The $L_2$ norm of the graphon loss propels the value of $p$ towards zero. Consequently, our optimization objective will eventually converge to the minimal stable features. In addition to the theoretical evidence provided thus far, we supplement our assertions with empirical support in Section \ref{sec:minimal-stable-feature} and \ref{sec:exp:str}.

\begin{table}[t]
\centering
\caption{\centering Statistics of datasets used in experiments.}
% \vspace{-2mm}
\label{table:dataset}
\begin{tabular}{lcrrrr}
\toprule
Dataset & Domain & \#Graphs & \#Nodes & \#Edges & \#Classes \\ 
\midrule
SYN-$b$     & base  & 8,000  & 230$\sim$247 & 542$\sim$1000 & 4 \\
Motif       & base  & 24,000 & 16.65 & 44.73    & 3  \\
Motif       & size  & 24,000 & 28.49 & 76.01    & 3  \\
CMNIST      & color & 56,000 & 75.00 & 1392.98  & 10 \\
Molhiv      & scaffold & 32,903 & 25.27 & 54.40  & 2 \\
Molhiv      & size     & 32,903 & 24.93 & 53.63  & 2 \\
Molbbbp     & scaffold & 2,039  & 24.06 & 51.90  & 2 \\
Molbbbp     & size     & 2,039  & 24.06 & 51.90  & 2 \\
\bottomrule
\end{tabular}
% \vspace{-4mm}
\end{table}

\section{Implementation Details}\label{apd:details}
\subsection{Datasets and Metrics}
We use SYN \cite{sui2021causal}, GOOD \cite{gui2022good} and OGB \cite{hu2020open} datasets, including Motif, CMNIST, Molhiv and Molbbbp, which are benchmarks for graph OOD tasks. For SYN-$b$, Motif and CMNIST, we use classification accuracy.
For Molhiv and Molbbbp, we follow studies \cite{hu2020open, gui2022good} and adopt ROC-AUC as the metric.
For all experimental results, we report the mean and standard deviation by conducting 10 random runs.
The statistical information of the dataset is shown in Table \ref{table:dataset}.

\subsection{Baselines}
To demonstrate the effectiveness of the proposed UIL, we choose the following baseline models and algorithms.

\begin{itemize}[leftmargin=4mm]
% \item \textbf{Graph Classification Models:}
% SortPool \cite{zhang2018end}, DiffPool \cite{ying2018hierarchical}, Top-$k$ Pool \cite{gao2019graph}, SAGPool \cite{lee2019self}, GCN \cite{kipf2016semi}, GIN \cite{xu2018how}, GAT \cite{velivckovic2018graph}, GATv2 \cite{brody2021attentive}, SuperGAT \cite{kim2020find}, GlobalAttention \cite{li2015gated}, AGNN \cite{thekumparampil2018attention}. These are popular GNNs for graph classification.

\item \textbf{General Generalization Algorithms:} 
ERM, IRM \cite{arjovsky2019invariant}, GroupDRO \cite{sagawa2019distributionally}, VREx \cite{krueger2021out}.

\item \textbf{Graph Generalization Algorithms:} 
DIR \cite{DIR}, GSAT \cite{miao2022interpretable}, CAL \cite{sui2021causal},  OOD-GNN \cite{li2021ood}, StableGNN \cite{fan2021generalizing}, CIGA \cite{chenlearning}, DisC \cite{fan2022debiasing}.

\item \textbf{Graph Data Augmentation:} 
DropEdge \cite{rong2019dropedge}, FLAG \cite{kong2022robust}, M-Mixup \cite{wang2021mixup}, $\Set{G}$-Mixup \cite{han2022g}, GREA \cite{liu2022graph}. 
% We provide more experimental details on these baseline methods in Appendix \ref{apd:baseline}.
\end{itemize}

\subsection{Experimental Settings}

\textbf{Training Settings.}\label{apd:training}
We use the NVIDIA GeForce RTX 3090 (24GB GPU) to conduct experiments.
To configure our model, we tune the hyperparameters in the following ranges: $\alpha$ and $\beta \in \{0.1, ..., 1.5\}$; $\rho \in \{0.5, ..., 1.0\}$; $|\hat{\Set{E}}| \in \{2,3,4,5,6\}$; batch size $\in \{32, 64, 128, 256\}$; learning rate $\in$\{1e-2, 1e-3, 1e-4\}.
All the baseline methods also follow the same hyperparameter search space as ours for parameter tuning to ensure that the comparison results are fair.
For ERM, GroupDRO \cite{sagawa2019distributionally}, IRM \cite{arjovsky2019invariant},  VREx \cite{krueger2021out}, M-Mixup \cite{wang2021mixup} and DIR \cite{DIR}, we copy the results from the original tables in the study \cite{gui2022good} and reproduce the results on Molbbbp dataset.
For CAL \cite{sui2021causal}, GSAT \cite{miao2022interpretable}, DropEdge \cite{rong2019dropedge}, GREA \cite{liu2022graph}, FLAG \cite{kong2022robust}, $\Set{G}$-Mixup \cite{han2022g}, CIGA \cite{chenlearning} and DisC \cite{fan2022debiasing}, they release the source codes.
Hence, we use their default model configurations from their original papers or codes to reproduce the results.
To help readers reproduce our results, we also provide an anonymous code link of our work.
All detailed hyperparameters are also provided in our codes.
% \zeminC{our model的具体参数设置最好也写一下？(空间不太够了，我在上面写了我们的超参搜索范围，应该可以吧。)}

\section{Additional Analyses}\label{apd:ana}
\subsection{Time Complexity}
Firstly, we define the average numbers of nodes and edges per graph as $|V|$ and $|E|$, respectively. Let $B$ denote the batch size for each training iteration, $l$ and $l_f$ denote the numbers of layers in the GNN backbone and feature estimator, respectively. $d$ and $d_f$ represent the dimensions of the hidden layers in the GNN backbone and feature estimator, respectively. For the stable feature learning objective, the time complexity is $\mathcal{O}(B(l_f|E|d_f + l|E|d))$. For the semantic invariant learning, the time complexity is $\mathcal{O}(B^2d)$. For the graphon estimation, the time complexity is $\mathcal{O}(B(|V|+|E|))$. For the structural invariant learning, the time complexity is $\mathcal{O}(C|\mathcal{\hat{E}}|N^2)$. For the regularization term, the time complexity is $\mathcal{O}(B(|V|+|E|))$. For simplicity, we assume $l=l_f$ and $d=d_f$. Hence, the time complexity of UIL is $\mathcal{O}(2Bl|E|d+B^2d+2B(|V|+|E|)+CN^2|\mathcal{\hat{E}}|)$.

\begin{table}[t]
\centering
\caption{\centering Comparison of running time and model size.}
\vspace{-2mm}
\label{table:running}
\resizebox{0.48\textwidth}{!}{\begin{tabular}{lcccccc}
\toprule
\multirow{3}{*}{Dataset} & \multicolumn{2}{c}{Baseline} & \multicolumn{2}{c}{CAL} & \multicolumn{2}{c}{Ours} \\ 
\cmidrule(r){2-3} \cmidrule(r){4-5} \cmidrule(r){6-7}
 & \makecell[c]{Running \\ time} & \makecell[c]{Model \\ size} & \makecell[c]{Running \\ time} & \makecell[c]{Model \\ size} & \makecell[c]{Running \\ time} & \makecell[c]{Model \\ size} \\ 
\midrule
Motif   & 51m 19s  & 1.515M & 1h 37m 15s & 2.213M    & 1h 33m 56s & 2.134M  \\
Molhiv  & 27m 20s  & 1.515M & 46m 14s    & 2.213M    & 43m 06s & 2.134M  \\
\bottomrule
\end{tabular}}
\vspace{-2mm}
\end{table}

\subsection{Running Time}
Compared to the original GNN model, UIL introduces additional complexities through a feature estimator, graphon estimation, and structural invariant learning. Despite these enhancements, the time and space costs remain within acceptable limits. We conducted empirical comparisons of the running time and model size between UIL, CAL, and the original model across various datasets. The results are shown in Table \ref{table:running}. Our analysis indicates that the UIL running time is approximately 1.5$\sim$2 times longer than that of the base model, and the model size is roughly 1.5 times larger. Despite these increases, our method competes closely with the current SOTA methods, such as CAL, in both running time and model size. Considering the significant performance improvements achieved by UIL, we argue that this model offers an advantageous performance-complexity trade-off. In practical applications, the additional computational complexity introduced by UIL is justifiable and manageable.

\section{Additional Experiments}\label{apd:exp}

\begin{table}[t]
% \vspace{-2mm}
\centering
\caption{Cut distances between graphons over different environments on SYN5 dataset.}
\vspace{-1mm}
\label{table:cut-syn5}
\resizebox{0.48\textwidth}{!}{\begin{tabular}{lccccc}
\toprule
Label & Cycle & House & Grid & Diamond & Avg. \\
\toprule
% \hline
$d_{full}$ & 13.33\scriptsize{$\pm$0.98} & 13.37\scriptsize{$\pm$0.87} & 13.41\scriptsize{$\pm$1.40} & 14.16\scriptsize{$\pm$1.15} & 13.57 \\
$d_{st}$ & 5.55\scriptsize{$\pm$2.42} & 4.52\scriptsize{$\pm$2.95} & 4.68\scriptsize{$\pm$3.45} & 6.96\scriptsize{$\pm$3.39} & 5.43 \\
$d_{en}$ & 13.15\scriptsize{$\pm$0.99} & 13.02\scriptsize{$\pm$0.86} & 13.00\scriptsize{$\pm$1.55} & 13.21\scriptsize{$\pm$1.21} & 13.10 \\
\bottomrule
\end{tabular}}
% \vspace{-4mm}
\end{table}

% \begin{table*}[t]
% \centering
% \caption{\centering Cut distances between full graphons and stable graphons over different environments on CMNIST dataset.}
% \vspace{-1mm}
% \label{table:cut-cmnist}
% \resizebox{1\textwidth}{!}{\begin{tabular}{lcccccccccc}
% \toprule
% Label & 0 & 1 & 2 & 3 & 4 & 5 & 6 & 7 & 8 & 9  \\
% \toprule
% % \hline
% $d_{full}$   & 11.19\scriptsize{$\pm$0.11}  & 11.06\scriptsize{$\pm$0.11}  & 11.28\scriptsize{$\pm$0.10} & 11.27\scriptsize{$\pm$0.15} & 11.20\scriptsize{$\pm$0.16} & 11.24\scriptsize{$\pm$0.15} & 11.17\scriptsize{$\pm$0.20} & 11.26\scriptsize{$\pm$0.10} & 11.22\scriptsize{$\pm$0.11} & 11.15\scriptsize{$\pm$0.11} \\
% $d_{st}$ & 6.89\scriptsize{$\pm$0.17} & 7.05\scriptsize{$\pm$0.09} & 6.89\scriptsize{$\pm$0.10} & 7.08\scriptsize{$\pm$0.14} & 6.99\scriptsize{$\pm$0.10} & 7.05\scriptsize{$\pm$0.10} & 7.08\scriptsize{$\pm$0.16} & 7.00\scriptsize{$\pm$0.15} & 7.00\scriptsize{$\pm$0.12} & 7.06\scriptsize{$\pm$0.10} \\
% $d_{en}$   & 10.89\scriptsize{$\pm$0.25}  & 10.56\scriptsize{$\pm$0.23}  & 10.79\scriptsize{$\pm$0.21} & 10.55\scriptsize{$\pm$0.41} & 10.37\scriptsize{$\pm$0.33} & 10.98\scriptsize{$\pm$0.85} & 11.01\scriptsize{$\pm$0.37} & 10.99\scriptsize{$\pm$0.32} & 10.76\scriptsize{$\pm$0.43} & 10.69\scriptsize{$\pm$0.36} \\
% \bottomrule
% \end{tabular}}
% \vspace{-1mm}
% \end{table*}

\subsection{Structural Invariance across Diverse Datasets}\label{apd:struct}
We curate a new dataset, SYN5, which mirrors SYN-$b$ except for the number of stable parts (\ie \textit{house, cycle, grid, diamond}). Each graph instance comprises an environmental subgraph containing $N_s$ stable parts, where $N_s$ is randomly sampled from the range 1$\sim$5. These stable parts collectively form the stable features of a graph instance.
We proceed to randomly sample two groups of graphs with identical labels $y$ but differing environments $e$ and $e'$, denoted as $\mathcal{G}^e$ and $\mathcal{G}^{e'}$. From the dataset, we can extract the ground-truth stable feature for each graph and compile two sets: $\mathcal{G}^e_{st}$ and $\mathcal{G}^{e'}_{st}$.
Next, we estimate two complete graphons $\mathcal{G}^e\Rightarrow W^e$ and $\mathcal{G}^{e'}\Rightarrow W^{e'}$, two stable graphons $\mathcal{G}^e_{st}\Rightarrow W^e_{st}$ and $\mathcal{G}^{e'}_{st}\Rightarrow W^{e'}_{st}$, and two environmental graphons $\mathcal{G}^e_{en}\Rightarrow W^e_{en}$ and $\mathcal{G}^{e'}_{en}\Rightarrow W^{e'}_{en}$. We then define $d_{full} = \delta_{\Box}(W^{e}, W^{e'})$, $d_{st} = \delta_{\Box}(W^{e}_{st}, W^{e'}_{st})$, and $d_{en} = \delta_{\Box}(W^{e}_{en}, W^{e'}_{en})$.
We execute experiments on both SYN5 and CMNIST, with results documented in Table \ref{table:cut-syn5}. It is evident from these results that the distances between complete graphons or environmental graphons are significantly larger than the distances between stable graphons across varying environments. This phenomenon underscores the notion that a class of graphs shares class-specific knowledge that exhibits structural invariance.

\end{document}